\documentclass{article}

\usepackage{iclr2019_conference,times}

\usepackage{multibib}

\usepackage{hhline}

\usepackage{amsmath,amsfonts,bm}

\def\eqref#1{equation~\ref{#1}}
\def\1{\bm{1}}

\DeclareMathAlphabet{\mathsfit}{\encodingdefault}{\sfdefault}{m}{sl}
\SetMathAlphabet{\mathsfit}{bold}{\encodingdefault}{\sfdefault}{bx}{n}

\usepackage[utf8]{inputenc} % allow utf-8 input
\usepackage[T1]{fontenc}    % use 8-bit T1 fonts
\usepackage{hyperref}       % hyperlinks
\usepackage{url}            % simple URL typesetting
\usepackage{booktabs}       % professional-quality tables
\usepackage{amsfonts}       % blackboard math symbols
\usepackage{nicefrac}       % compact symbols for 1/2, etc.
\usepackage{microtype}      % microtypography
\usepackage{graphicx}
\usepackage{subfigure}
\usepackage{amsmath}
\usepackage{multirow}
\usepackage{amsthm}
\usepackage{amssymb}
\usepackage{mathtools}
\usepackage{color}
\usepackage{xcolor}
\usepackage{MnSymbol}
\usepackage{makecell}
\usepackage{arydshln}
\usepackage[shortlabels]{enumitem}
\usepackage{booktabs}
\usepackage{caption}
\usepackage{wrapfig,lipsum}

\usepackage{sidecap}

\newcommand*{\QEDA}{\hfill\ensuremath{\blacksquare}}%
\newcommand*{\QEDB}{\hfill\ensuremath{\square}}%
\newcommand\independent{\protect\mathpalette{\protect\independenT}{\perp}}
\def\independenT#1#2{\mathrel{\rlap{$#1#2$}\mkern2mu{#1#2}}}
\newcommand{\ourl}{Multimodal Factorization Model}
\newcommand{\ours}{MFM}

\newtheorem{proposition}{Proposition}

\newtheorem{lemma}{Lemma}

\makeatletter
   \def\tagform@#1{\maketag@@@{\normalsize(#1)\@@italiccorr}} 
\makeatother
\iclrfinalcopy
\title{\Large Learning Factorized Multimodal Representations}

\author{Yao-Hung Hubert Tsai$^{*1}$, $\ $ Paul Pu Liang\thanks{equal contributions}$\,\,\,^1$, \\
  {\bf Amir Zadeh}$^{2}${\bf,} $\ $ {\bf Louis-Philippe Morency}$^{2}${\bf,} $\ $ {\bf Ruslan Salakhutdinov}$^{1}$ \\
  \{$^1$Machine Learning Department, $^2$Language Technologies Institute\}, Carnegie Mellon University\\
  \texttt{\{yaohungt,pliang,abagherz,morency,rsalakhu\}@cs.cmu.edu} \\
}

\begin{document}
\vspace{-2mm}
\maketitle
\vspace{-2mm}

\begin{abstract}
Learning multimodal representations is a fundamentally complex research problem due to the presence of multiple heterogeneous sources of information. Although the presence of multiple modalities provides additional valuable information, there are two key challenges to address when learning from multimodal data: 1) models must learn the complex intra-modal and cross-modal interactions for prediction and 2) models must be robust to unexpected missing or noisy modalities during testing. In this paper, we propose to optimize for a joint generative-discriminative objective across multimodal data and labels. We introduce a model that factorizes representations into two sets of independent factors: \textit{multimodal discriminative} and \textit{modality-specific generative} factors. Multimodal discriminative factors are shared across all modalities and contain joint multimodal features required for discriminative tasks such as sentiment prediction. Modality-specific generative factors are unique for each modality and contain the information required for generating data. Experimental results show that our model is able to learn meaningful multimodal representations that achieve state-of-the-art or competitive performance on six multimodal datasets. Our model demonstrates flexible generative capabilities by conditioning on independent factors and can reconstruct missing modalities without significantly impacting performance. Lastly, we interpret our factorized representations to understand the interactions that influence multimodal learning.
\end{abstract}

%!TEX root = ../nips_2018.tex

\vspace{-3mm}
\section{Introduction}
\label{sec:intro}

Multimodal machine learning involves learning from data across multiple modalities~\citep{baltruvsaitis2017multimodal}. It is a challenging yet crucial research area with real-world applications in robotics~\citep{8039024}, dialogue systems~\citep{Pittermann2010}, intelligent tutoring systems~\citep{PETROVICA2017437}, and healthcare diagnosis~\citep{5373931}. At the heart of many multimodal modeling tasks lies the challenge of learning rich representations from multiple modalities. For example, analyzing multimedia content requires learning multimodal representations across the language, visual, and acoustic modalities~\citep{DBLP:journals/corr/ChoCB15}. Although the presence of multiple modalities provides additional valuable information, there are two key challenges to address when learning from multimodal data: 1) models must learn the complex intra-modal and cross-modal interactions for prediction~\citep{tensoremnlp17}, and 2) trained models must be robust to unexpected missing or noisy modalities during testing~\citep{ngiam2011multimodal}.

In this paper, we propose to optimize for a joint generative-discriminative objective across multimodal data and labels. The discriminative objective ensures that the representations learned are rich in intra-modal and cross-modal features useful towards predicting the label, while the generative objective allows the model to infer missing modalities at test time and deal with the presence of noisy modalities. To this end, we introduce the \ourl \ (\ours \ in Figure \ref{fig:illus}) that factorizes multimodal representations into \textit{multimodal discriminative} factors and \textit{modality-specific generative} factors. Multimodal discriminative factors are shared across all modalities and contain joint multimodal features required for discriminative tasks. Modality-specific generative factors are unique for each modality and contain the information required for generating each modality. We believe that factorizing multimodal representations into different explanatory factors can help each factor focus on learning from a subset of the joint information across multimodal data and labels. This method is in contrast to jointly learning a single factor that summarizes all generative and discriminative information~\citep{srivastava2012multimodal}. To sum up, \ours \ defines a joint distribution over multimodal data, and by the conditional independence assumptions in the assumed graphical model, both generative and discriminative aspects are taken into account. Our model design further provides interpretability of the factorized representations.

Through an extensive set of experiments, we show that \ours \ learns improved multimodal representations with these characteristics: 1) The multimodal discriminative factors achieve state-of-the-art or competitive performance on six multimodal time series datasets. We also demonstrate that \ours \ can generalize by integrating it with other existing multimodal discriminative models. 2) \ours \ allows flexible generation concerning multimodal discriminative factors (labels) and modality-specific generative factors (styles). We further show that we can perform reconstruction of missing modalities from observed modalities without significantly impacting discriminative performance. Finally, we interpret our learned representations using information-based and gradient-based methods, allowing us to understand the contributions of individual factors towards multimodal prediction and generation.
%!TEX root = ../nips_2018.tex

\section{\ourl }

\begin{figure}[t!]
\centering
\vspace{-8mm}
\includegraphics[width=0.9\textwidth]{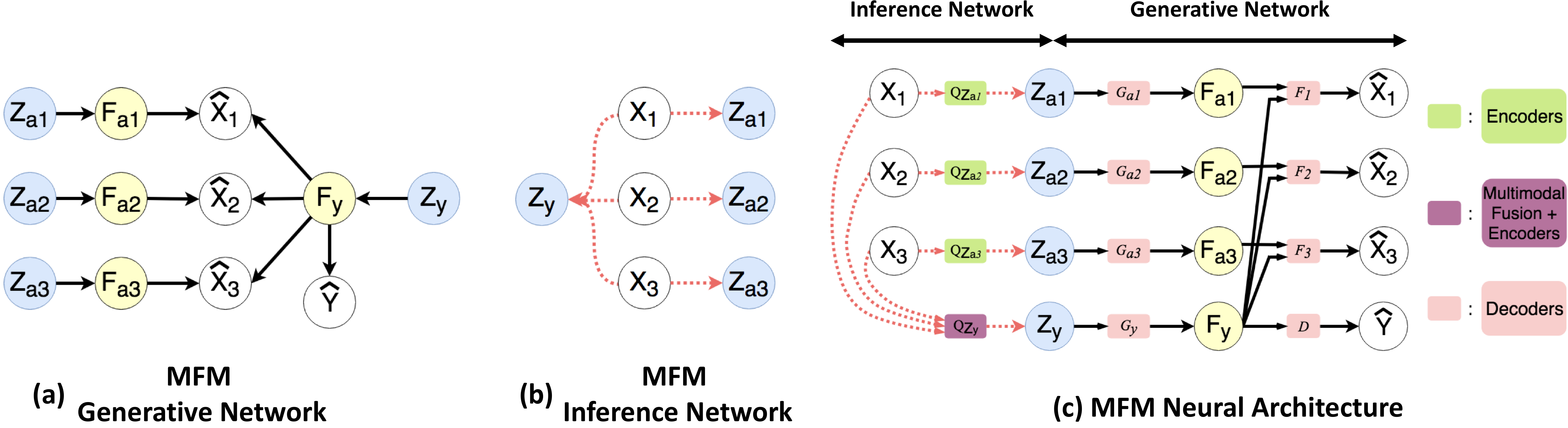}
\caption{\small Illustration of the proposed \ourl \ (\ours) with three modalities. \ours \ factorizes multimodal representations into {\em multimodal discriminative} factors $\mathbf{F_y}$ and {\em modality-specific generative} factors $\mathbf{F_a}_{\{1:M\}}$. (a) \ours \ Generative Network with latent variables $\{ \mathbf{Z_y}, \mathbf{Z_a}_{\{1:M\}} \}$, factors $\{ \mathbf{F_y}, \mathbf{F_a}_{\{1:M\}} \}$, generated multimodal data $\mathbf{\hat{X}}_{1:3}$ and labels $\mathbf{\hat{Y}}$. (b) \ours \ Inference Network. (c) \ours \ Neural Architecture. Best viewed zoomed in and in color.\vspace{-5mm}}
\label{fig:illus}
\end{figure}

\ourl \ (MFM) is a latent variable model (Figure \ref{fig:illus}(a)) with conditional independence assumptions over multimodal discriminative factors and modality-specific generative factors. According to these assumptions, we propose a factorization over the joint distribution of multimodal data (Section \ref{subsec:disen}). Since exact posterior inference on this factorized distribution can be intractable, we propose an approximate inference algorithm based on minimizing a joint-distribution Wasserstein distance over multimodal data (Section \ref{subsec:wae}). Finally, we derive the \ours \ objective by approximating the joint-distribution Wasserstein distance via a generalized mean-field assumption.

\textbf{Notation:} We define $\mathbf{X}_{1:M}$ as the multimodal data from $M$ modalities and $\mathbf{Y}$ as the labels, with joint distribution ${P}_{\mathbf{X}_{1:M},\mathbf{Y}} = {P}(\mathbf{X}_{1:M},\mathbf{Y})$. Let $\mathbf{\hat{X}}_{1:M}$ denote the generated multimodal data and $\mathbf{\hat{Y}}$ denote the generated labels, with joint distribution ${P}_{\mathbf{\hat{X}}_{1:M},\mathbf{\hat{Y}}} = {P}(\mathbf{\hat{X}}_{1:M},\mathbf{\hat{Y}})$. 
\vspace{-3mm}

\subsection{Factorized Multimodal Representations}
\label{subsec:disen}
\vspace{-2mm}
To factorize multimodal representations into multimodal discriminative factors and modality-specific generative factors, \ours \ assumes a Bayesian network structure as shown in Figure~\ref{fig:illus}(a). In this graphical model, factors $\mathbf{F_y}$ and $\mathbf{F_a}_{\{1:M\}}$ are generated from mutually independent latent variables $\mathbf{Z} = [\mathbf{Z_y}, \mathbf{Z_a}_{\{1:M\}}]$ with prior $P_{\mathbf{Z}}$. In particular, $\mathbf{Z_y}$ generates the multimodal discriminative factor $\mathbf{F_y}$ and $\mathbf{Z_a}_{\{1:M\}}$ generate modality-specific generative factors $\mathbf{F_a}_{\{1:M\}}$. By  construction, $\mathbf{F_y}$ contributes to the generation of $\mathbf{\hat{Y}}$ while $\{ \mathbf{F_y}, \mathbf{F_a}_{i} \}$ both contribute to the generation of $\mathbf{\hat{X}}_{i}$. As a result, the joint distribution ${P}(\mathbf{\hat{X}}_{1:M},\mathbf{\hat{Y}})$ can be factorized as follows:
\begin{equation}
\label{eq:factor}
\scriptsize
\begin{split}
&{P}(\mathbf{\hat{X}}_{1:M},\mathbf{\hat{Y}}) = \int_{\mathbf{F},\mathbf{Z}} {P}(\mathbf{\hat{X}}_{1:M},\mathbf{\hat{Y}}|\mathbf{F}) P(\mathbf{F}|\mathbf{Z}) P(\mathbf{Z}) d \mathbf{F} d \mathbf{Z}\\
=\,\,& \int_{\substack{\mathbf{F_y}, \mathbf{F_a}_{\{1:M\}} \\ \mathbf{Z_y}, \mathbf{Z_a}_{\{1:M\}}}} \Big(P(\mathbf{\hat{Y}|F_y}) \prod_{i=1}^M{P}(\mathbf{\hat{X}}_{i}|\mathbf{F_a}_{i},\mathbf{F_y})\Big) \Big( P(\mathbf{F_y}|\mathbf{Z_y}) \prod_{i=1}^M{P}(\mathbf{F_a}_{i}|\mathbf{Z_a}_{i}) \Big) \Big( P(\mathbf{Z_y}) \prod_{i=1}^M{P}(\mathbf{Z_a}_{i}) \Big) d \mathbf{F} d \mathbf{Z},  
\end{split}
\end{equation}
with $d \mathbf{F} = d \mathbf{F_y} \prod_{i=1}^M d\mathbf{F_a}_{i} $ and $d \mathbf{Z} = d \mathbf{Z_y} \prod_{i=1}^M d\mathbf{Z_a}_{i}$.

\noindent Exact posterior inference in Equation~\ref{eq:factor} may be analytically intractable due to the integration over $\mathbf{Z}$. We therefore resort to using an approximate inference distribution $Q(\mathbf{Z}|\mathbf{X}_{1:M},\mathbf{Y})$ as detailed in the following subsection. As a result, \ours \ can be viewed as an autoencoding structure that consists of encoder (inference) and decoder (generative) modules (Figure~\ref{fig:illus}(c)). The encoder module for $Q(\cdot|\cdot)$ allows us to easily sample $\mathbf{Z}$ from an approximate posterior. The decoder modules are parametrized according to the factorization of ${P}(\mathbf{\hat{X}}_{1:M},\mathbf{\hat{Y}}|\mathbf{Z})$ as given by Equation~\ref{eq:factor} and Figure~\ref{fig:illus}(a).

\subsection{Minimizing Joint-Distribution Wasserstein Distance over Multimodal Data}
\label{subsec:wae}
\vspace{-1mm}

Two common choices for approximate inference in autoencoding structures are Variational Autoencoders (VAEs)~\citep{kingma2013auto} and Wasserstein Autoencoders (WAEs)~\citep{zhao2017infovae,tolstikhin2017wasserstein}. The former optimizes the evidence lower bound objective (ELBO), and the latter derives an approximation for the primal form of the Wasserstein distance. We consider the latter since it simultaneously results in better latent factor disentanglement~\citep{zhao2017infovae,rubenstein2018latent} and better sample generation quality than its counterparts~\citep{chen2016infogan,higgins2016beta,kingma2013auto}. However, WAEs are designed for unimodal data and do not consider factorized distributions over latent variables that generate multimodal data. Therefore, we propose a variant for handling factorized joint distributions over multimodal data.

As suggested by~\cite{kingma2013auto}, we adopt the design of nonlinear mappings (i.e. neural network architectures) in the encoder and decoder (Figure~\ref{fig:illus} (c)). For the encoder $Q(\mathbf{Z}|\mathbf{X}_{1:M},\mathbf{Y})$, we learn a deterministic mapping $Q_{enc}:\mathbf{X}_{1:M},\mathbf{Y} \rightarrow \mathbf{Z}$~\citep{rubenstein2018latent,tolstikhin2017wasserstein}. For the decoder, we define the generation process from latent variables as $G_y: \mathbf{Z_y} \rightarrow \mathbf{F_y}$, $G_{a\{1:M\}}: \mathbf{Z}_{\mathbf{a}\{1:M\}} \rightarrow \mathbf{F}_{\mathbf{a}\{1:M\}}$, $D: \mathbf{F_y} \rightarrow \mathbf{\hat{Y}}$, and $F_{1:M}: \mathbf{F_y}, \mathbf{F}_{\mathbf{a}\{1:M\}} \rightarrow \mathbf{\hat{X}}_{1:M}$, where $G_{y}, G_{a\{1:M\}}, D$ and $F_{1:M}$ are deterministic functions parametrized by neural networks.

Let $W_c({P}_{\mathbf{X}_{1:M},\mathbf{Y}}, {P}_{\mathbf{\hat{X}}_{1:M},\mathbf{\hat{Y}}})$ denote the joint-distribution Wasserstein distance over multimodal data under cost function ${c_X}_{i}$ and $c_Y$. We choose the squared cost $c(a, b) = \|a - b\|_2^2$, allowing us to minimize the 2-Wasserstein distance. The cost function can be defined not only on static data but also on time series data such as text, audio and videos. For example, given time series data $\mathbf{X} = [X^1, X^2, \cdots , X^T ]$ and $\mathbf{\hat{X}} = [\hat{X}^1, \hat{X}^2, \cdots , \hat{X}^T ]$, we define $c(\mathbf{X}, \mathbf{\hat{X}}) = \sum_{t=1}^{T} \|X^t - \hat{X}^t \|_2^2$.

With conditional independence assumptions in Equation~\ref{eq:factor}, we express $W_c({P}_{\mathbf{X}_{1:M},\mathbf{Y}}, {P}_{\mathbf{\hat{X}}_{1:M},\mathbf{\hat{Y}}})$ as:
%!TEX root = ../nips_2018.tex

\begin{proposition}
\label{prop:multi}
For any functions $G_y: \mathbf{Z_y} \rightarrow \mathbf{F_y}$, $G_{a\{1:M\}}: \mathbf{Z}_{\mathbf{a}\{1:M\}} \rightarrow \mathbf{F}_{\mathbf{a}\{1:M\}}$, $D: \mathbf{F_y} \rightarrow \mathbf{\hat{Y}}$, and $F_{1:M}: \mathbf{F}_{\mathbf{a}\{1:M\}}, \mathbf{F_y} \rightarrow \mathbf{\hat{X}}_{1:M}$, we have $W_c({P}_{\mathbf{X}_{1:M},\mathbf{Y}}, {P}_{\mathbf{\hat{X}}_{1:M},\mathbf{\hat{Y}}}) = $
\begin{equation}
\small
\label{eq:propmulti}
\begin{split}
&\underset{{Q_{\mathbf{Z}}} = {P_{\mathbf{Z}}}}{\mathrm{inf}}\,\,\mathbf{E}_{{P}_{\mathbf{X}_{1:M},\mathbf{Y}}}\mathbf{E}_{{Q}({\mathbf{Z}|\mathbf{X}_{1:M},\mathbf{Y}})}\Bigg[\sum_{i=1}^{M}c_{X_i}\Big(\mathbf{X}_i, F_i\big(G_{ai}(\mathbf{Z}_{\mathbf{a}i}), G_y(\mathbf{Z_y})\big)\Big) + c_Y\Big(\mathbf{Y}, D\big(G_y(\mathbf{Z_y})\big)\Big)\Bigg],
\end{split}
\end{equation}
where ${P_{\mathbf{Z}}}$ is the prior over $\mathbf{Z} = [\mathbf{Z_y}, \mathbf{Z_a}_{\{1,M\}} ]$ and ${Q_{\mathbf{Z}}}$ is the aggregated posterior of the proposed approximate inference distribution ${Q}({\mathbf{Z}|\mathbf{X}_{1:M},\mathbf{Y}})$.
\end{proposition}

\vspace{-1mm}

{\em Proof:} The proof is adapted from Tolstikhin {\em et al.} \citep{tolstikhin2017wasserstein}. The two differences are: (1) we show that $P(\mathbf{\hat{X}}_{1:M}, \mathbf{\hat{Y}}|\mathbf{Z} = z)$ are Dirac for all $z \in \mathcal{Z}$, and (2) we use the fact that $c ((\mathbf{X}_{1:M}, \mathbf{Y}), (\mathbf{\hat{X}}_{1:M}, \mathbf{\hat{Y}})) = \sum_{i=1}^M {c_{X}}_i (\mathbf{X}_i, \mathbf{\hat{X}}_i) + c_{Y} (\mathbf{Y}, \mathbf{\hat{Y}})$. Please refer to the supplementary material for proof details. \QEDA
\vspace{-1mm}

The constraint on $Q_\mathbf{Z} = P_\mathbf{Z}$ in Proposition~\ref{prop:multi} is hard to satisfy. To obtain a numerical solution, we first relax the constraint by performing a generalized mean field assumption on $Q$ according to the conditional independence as shown in the inference network of Figure~\ref{fig:illus} (b): 
\vspace{-2mm}
\begin{equation}
\small
\label{eq:mean}
Q(\mathbf{Z|X}_{1:M},\mathbf{Y}) \coloneqq Q(\mathbf{Z|X}_{1:M}) \coloneqq Q({\mathbf{Z_y}|\mathbf{X}_{1:M}}) \prod_{i=1}^M Q(\mathbf{Z_{a}}_i|\mathbf{X}_i).
\vspace{-2mm}
\end{equation}
The intuition here is based on our design that $\mathbf{Z_y}$ generates the multimodal discriminative factor $\mathbf{F_y}$ and $\mathbf{Z_a}_{\{1:M\}}$ generate modality-specific generative factors $\mathbf{F_a}_{\{1:M\}}$. Therefore, the inference for $\mathbf{Z_y}$ should depend on all modalities $\mathbf{X}_{1:M}$ and the inference for $\mathbf{Z_a}_i$ should depend only on the specific modality $\mathbf{X}_i$. Following this assumption, we define $\mathcal{Q}$ as a nonparametric set of all encoders that fulfill the factorization in Equation~\ref{eq:mean}. A penalty term is added into our objective to find the $Q(\mathbf{Z} | \cdot) \in \mathcal{Q}$ that is the closest to prior $P_\mathbf{Z}$, thereby approximately enforcing the constraint $Q_\mathbf{Z} = P_\mathbf{Z}$:
\begin{equation}
\vspace{-1mm}
\label{eq:approxmulti}
\small
\begin{split}
&\underset{F, G_{a\{1:M\}}, G_y, D}{\mathrm{min}}\,\,\underset{{Q(\mathbf{Z}|\cdot)\in \mathcal{Q}}}{\mathrm{inf}}\,\,\mathbf{E}_{{P}_{\mathbf{X}_{1:M},\mathbf{Y}}}\mathbf{E}_{{Q}({\mathbf{Z}_{\mathbf{a}1}|\mathbf{X}_{1}})} \cdots \mathbf{E}_{{Q}({\mathbf{Z}_{\mathbf{a}M}|\mathbf{X}_{M}})} \mathbf{E}_{{Q}({\mathbf{Z_y}|\mathbf{X}_{1:M}})}\\
&\Bigg[\sum_{i=1}^M c_{X_i}\Big(\mathbf{X}_i, F\big(G_{ai}(\mathbf{Z}_{\mathbf{a}i}), G_y(\mathbf{Z_y})\big)\Big) + c_Y\Big(\mathbf{Y}, D\big(G_y(\mathbf{Z_y})\big)\Big)\Bigg]  + \lambda \mathcal{MMD}({Q}_{\mathbf{Z}}, {P}_{\mathbf{Z}}),
\end{split}
\vspace{-2mm}
\end{equation}
where $\lambda$ is a hyper-parameter and $\mathcal{MMD}$ is the Maximum Mean Discrepancy \citep{gretton2012kernel} as a divergence measure between ${Q}_\mathbf{Z}$ and ${P}_\mathbf{Z}$. The prior ${P}_\mathbf{Z}$ is chosen as a centered isotropic Gaussian $\mathcal{N}(\mathbf{0}, \mathbf{I})$, so that it implicitly enforces independence between the latent variables $\mathbf{Z} = [\mathbf{Z_y}, \mathbf{Z_a}_{\{1,M\}}]$~\citep{higgins2016beta,kingma2013auto,rubenstein2018latent}. 

Equation~\ref{eq:approxmulti} represents our hybrid generative-discriminative optimization objective over multimodal data: the first loss term $ \sum_{i=1}^M c_{X_i} (\mathbf{X}_i, F (G_{ai}(\mathbf{Z}_{\mathbf{a}i}), G_y(\mathbf{Z_y}))) $ is the generative objective based on reconstruction of multimodal data and the second term $ c_Y(\mathbf{Y}, D(G_y(\mathbf{Z_y}))) $ is the discriminative objective. In practice we compute the expectations in Equation~\ref{eq:approxmulti} using empirical estimates over the training data. The neural architecture of \ours \ is illustrated in Figure~\ref{fig:illus}(c).

\vspace{-1mm}
\subsection{Surrogate Inference for Missing Modalities}
\label{sec:missing}

A key challenge in multimodal learning involves dealing with missing modalities. A good multimodal model should be able to infer the missing modality conditioned on the observed modalities and perform predictions based only on the observed modalities. To achieve this objective, the inference process of \ours \ can be easily adapted using a surrogate inference network to reconstruct the missing modality given the observed modalities. Formally, let $\Phi$ denote the surrogate inference network. The generation of missing modality $\hat{\mathbf{X}}_1$ given the observed modalities $\mathbf{X}_{2:M}$ can be formulated as 
\begin{equation}
\vspace{-1mm}
\begin{split}
\scriptsize
&\Phi^* = \underset{\Phi}{\mathrm{argmin}}\,\mathbf{E}_{P_{\mathbf{X}_{2:M}, \hat{\mathbf{X}}_1}}\, \Big( -\mathrm{log}\,P_{\Phi}(\hat{\mathbf{X}}_1|\mathbf{X}_{2:M}) \Big)\\
\,\,\mathrm{with}\,\,&P_{\Phi}(\hat{\mathbf{X}}_1|\mathbf{X}_{2:M}) := \int P(\hat{\mathbf{X}}_1|\mathbf{Z}_{\mathbf{a}1}, \mathbf{Z_y})Q_\Phi(\mathbf{Z}_{\mathbf{a}1}|\mathbf{X}_{2:M})Q_\Phi(\mathbf{Z_y}|\mathbf{X}_{2:M}) d \mathbf{Z}_{\mathbf{a}1}d \mathbf{Z_y}.
\label{eq:Phi}
\end{split}
\vspace{-1mm}
\end{equation}
Similar to Section~\ref{subsec:wae}, we use deterministic mappings in $Q_\Phi(\cdot|\cdot)$ and $Q_\Phi(\mathbf{Z_y}|\cdot)$ is also used for prediction $P_{\mathrm{\Phi}}(\hat{\mathbf{Y}}|\mathbf{X}_{2:M}):= \int P(\hat{\mathbf{Y}}| \mathbf{Z_y})Q_\Phi(\mathbf{Z_y}|\mathbf{X}_{2:M}) d \mathbf{Z_y}$. Equation~\ref{eq:Phi} suggests that in the presence of missing modalities, we only need to infer the latent codes rather than the entire modality. 

\vspace{-1mm}
\subsection{Encoder and Decoder Design}
\label{enc_dec}

We now discuss the implementation choices for the \ours \ neural architecture in Figure~\ref{fig:illus}(c). The encoder $Q(\mathbf{Z_y}|\mathbf{X}_{1:M})$ can be parametrized by any model that performs multimodal fusion~\citep{morency2011towards,tensoremnlp17}. For multimodal image datasets, we adopt Convolutional Neural Networks (CNNs) and Fully-Connected Neural Networks (FCNNs) with late fusion~\citep{Nojavanasghari:2016:DMF:2993148.2993176} as our encoder $Q(\mathbf{Z_y}|\mathbf{X}_{1:M})$. The remaining functions in \ours \ are also parametrized by CNNs and FCNNs. For multimodal time series datasets, we choose the Memory Fusion Network (MFN)~\citep{zadeh2018memory} as our multimodal encoder $Q(\mathbf{Z_y}|\mathbf{X}_{1:M})$. We use Long Short-term Memory (LSTM) networks~\citep{hochreiter1997long} for functions $Q(\mathbf{Z_a}_{\{1:M\}}|\mathbf{X}_{1:M})$, decoder LSTM networks~\citep{cho-al-emnlp14} for functions $F_{1:M}$, and FCNNs for functions $G_y$, ${G_a}_{\{1:M\}}$ and $D$. Details are provided in the appendix and the code is available at \url{https://github.com/pliang279/factorized/}.
\vspace{-2mm}
\section{Experiments}
\label{sec:exp}
\vspace{-1mm}

\newcolumntype{C}[1]{>{\centering\let\newline\\\arraybackslash\hspace{0pt}}m{#1}}

\newcolumntype{K}[1]{>{\centering\arraybackslash}p{#1}}

\begin{figure}[t!]
\vspace{-8mm}
    \centering
    \begin{minipage}{0.35\textwidth}
        \centering
        \includegraphics[width=0.8\linewidth]{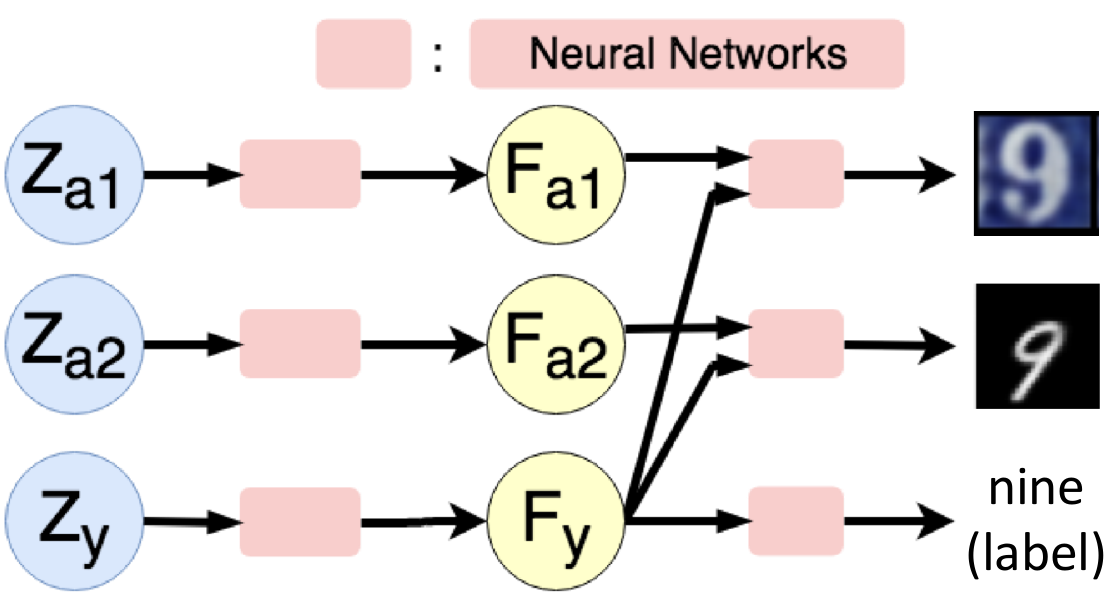}
        \captionsetup{labelformat=empty}
        \vspace{-2mm}
        \caption*{(a)\vspace{1mm}}
        %\label{(a)}
        \fontsize{6}{9}\selectfont
        \setlength\tabcolsep{1.2pt}
        \begin{tabular}{|K{0.7cm} || *{2}{K{1.2cm}} *{2}{K{0.7cm}}|}
        \hline
        	Method	 & UM(SVHN)  &  UM(MNIST) & MM & MFM \\
        	\hline \hline
        	Acc. & 91.84 & 99.01 & 99.20  & \textbf{99.36} \\
        \hline
        \end{tabular}
         \captionsetup{labelformat=empty}
         \vspace{-3mm}
        \caption*{(b) }
    \end{minipage}%
    \hspace{5mm}
    \begin{minipage}{0.5\textwidth}
        \centering
       \includegraphics[width=0.95\linewidth]{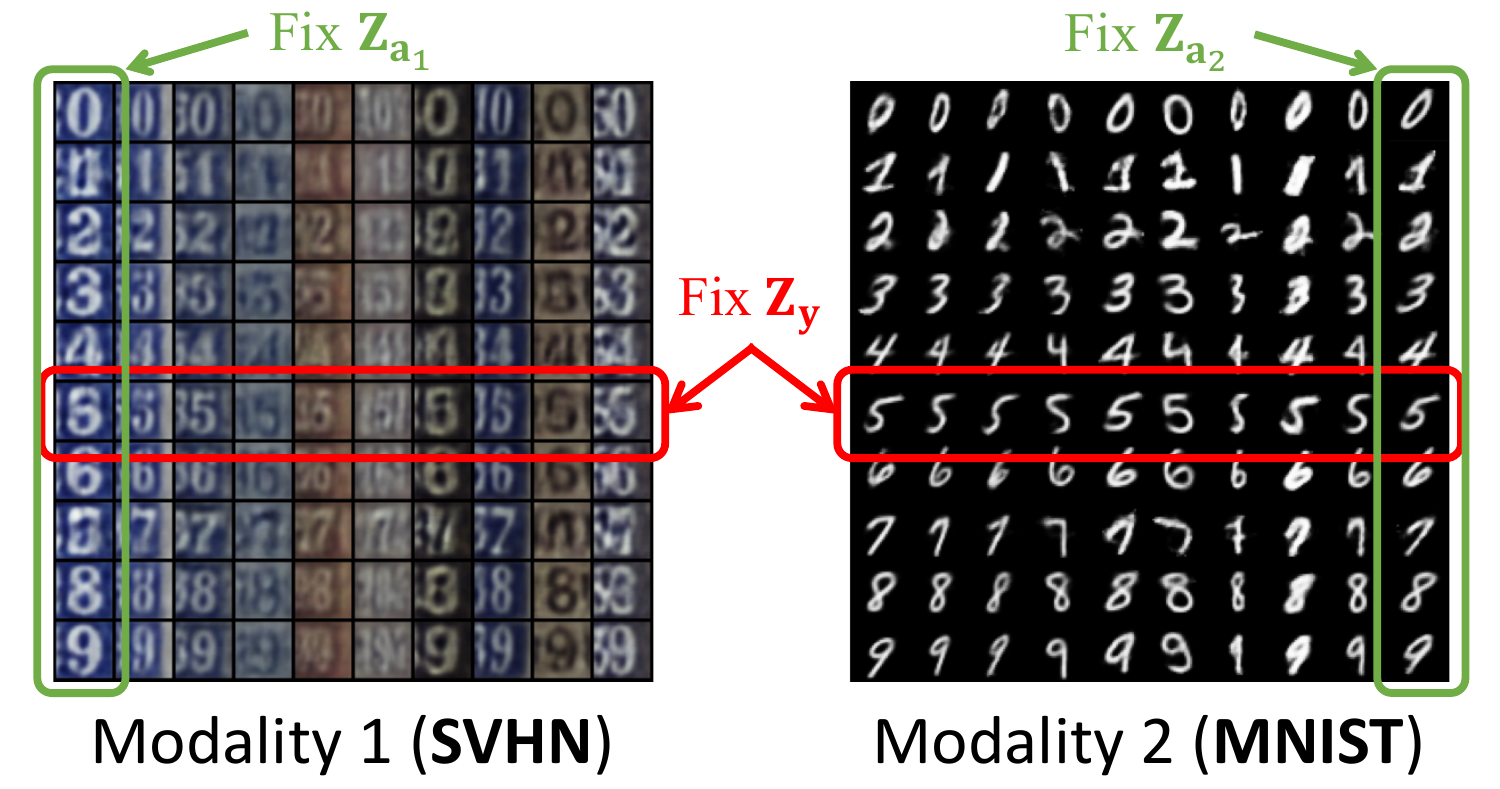}
       \vspace{-2mm}
         \captionsetup{labelformat=empty}
        \vspace{-1mm}
        \caption*{(c)}
        %\label{fig:prob1_6_1}
    \end{minipage}
\vspace{-4mm}
\caption{\small (a) \ours \ generative network for multimodal image dataset SVHN+MNIST, (b) unimodal and multimodal classification accuracies, and (c) conditional generation for SVHN and MNIST digits. \ours \ shows improved capabilities in digit prediction as well as flexible generation of both images based on labels and styles.
\vspace{-4mm}}
\label{fig:control}
\end{figure}

In order to show that \ours \ learns multimodal representations that are discriminative, generative and interpretable, we design the following experiments. We begin with a multimodal synthetic image dataset that allows us to examine whether \ours \ displays discriminative and generative capabilities from factorized latent variables. Utilizing image datasets allows us to clearly visualize the generative capabilities of \ours. We then transition to six more challenging real-world multimodal video datasets to 1) rigorously evaluate the discriminative capabilities of \ours \ in comparison with existing baselines, 2) analyze the importance of each design component through ablation studies, 3) assess the robustness of \ours's modality reconstruction and prediction capabilities to missing modalities, and 4) interpret the learned representations using information-based and gradient-based methods to understand the contributions of individual factors towards multimodal prediction and generation.

\subsection{Multimodal Synthetic Image Dataset}

In this section, we study \ours \ on a synthetic image dataset that considers SVHN~\citep{noauthororeditor} and MNIST~\citep{Lecun98gradient-basedlearning} as the two modalities. SVHN and MNIST are images with different styles but the same labels (digits $0\sim 9$). We randomly pair $100,000$ SVHN and MNIST images that have the same label, creating a multimodal dataset which we call SVHN+MNIST. $80,000$ pairs are used for training and the rest for testing. To justify that \ours \ is able to learn improved multimodal representations, we show both classification and generation results on SVHN+MNIST in Figure \ref{fig:control}.

\textbf{Prediction:} We perform experiments on both unimodal and multimodal classification tasks. UM denotes a unimodal baseline that performs prediction given only one modality as input and MM denotes a multimodal discriminative baseline that performs prediction given both images~\citep{Nojavanasghari:2016:DMF:2993148.2993176}. We compare the results for UM(SVHN), UM(MNIST), MM and \ours \ on SVHN+MNIST in Figure \ref{fig:control}(b). We achieve better classification performance from unimodal to multimodal which is not surprising since more information is given. More importantly, \ours \ outperforms MM, which suggests that \ours \ learns improved factorized representations for discriminative tasks.

\textbf{Generation:} We generate images using the \ours \ generative network (Figure \ref{fig:control}(a)). We fix one variable out of $\mathbf{Z} = [ \mathbf{Z_a}_1$, $\mathbf{Z_a}_2$, and $\mathbf{Z_y} ]$ and randomly sample the other two variables from prior $P_{\mathbf{Z}}$. From Figure \ref{fig:control}(c), we observe that \ours \ shows flexible generation of SVHN and MNIST images based on labels and styles. This suggests that \ours \ is able to factorize multimodal representations into multimodal discriminative factors (labels) and modality-specific generative factors (styles).

\vspace{-2mm}
\subsection{Multimodal Time Series Datasets}
\label{disc}
\vspace{-1mm}

\newcolumntype{K}[1]{>{\centering\arraybackslash}p{#1}}
%\definecolor{gg}{RGB}{45,130,45}
\definecolor{gg}{RGB}{0,0,0}
\begin{table*}[t!] %[!htbp]
\vspace{-5mm}
\caption{\small Results for multimodal speaker traits recognition on POM, multimodal sentiment analysis on CMU-MOSI, ICT-MMMO, YouTube, MOUD, and multimodal emotion recognition on IEMOCAP. SOTA1 and SOTA2 refer to the previous best and second best state-of-the-art respectively, and $\Delta_{SOTA}$ shows improvement over SOTA1. Symbols depict the baseline giving the result: $\#$ \textit{MFN}, $\ddagger$ \textit{MARN}, $\ast$ \textit{TFN}, $\dagger$ \textit{BC-LSTM}, $\diamond$ \textit{MV-LSTM}, $\S$ \textit{EF-LSTM}, $\flat$ \textit{DF}, $\heartsuit$ \textit{SVM}, $\bullet$ \textit{RF}. For detailed tables with results for all models, please refer to the appendix.}

\fontsize{7}{10}\selectfont
%\centering
\setlength\tabcolsep{0.7pt}
\vspace{-2mm}
\begin{tabular}{|c || *{16}{K{0.752cm}} |}
\hline
Dataset & \multicolumn{16}{c|}{{POM Personality Traits}} \\
Task & \multicolumn{1}{c}{Con} & \multicolumn{1}{c}{Pas} & \multicolumn{1}{c}{Voi} & \multicolumn{1}{c}{Dom} & \multicolumn{1}{c}{Cre} & \multicolumn{1}{c}{Viv} & \multicolumn{1}{c}{Exp} & \multicolumn{1}{c}{Ent} & \multicolumn{1}{c}{Res} & \multicolumn{1}{c}{Tru} & \multicolumn{1}{c}{Rel} & \multicolumn{1}{c}{Out} & \multicolumn{1}{c}{Tho} & \multicolumn{1}{c}{Ner} & \multicolumn{1}{c}{Per} & \multicolumn{1}{c|}{Hum}\\
Metric  & \multicolumn{16}{c|}{$r$} \\ 
\hline \hline
SOTA2 & 0.359$^\dagger$&0.425$^\dagger$&0.166$^\ddagger$&0.235$^\ddagger$&0.358$^\dagger$&0.417$^\dagger$&0.450$^\dagger$&0.378$^\ddagger$&0.295$^\diamond$&0.237$^\diamond$&0.215$^\ddagger$&0.238$^\diamond$&0.363$^\dagger$&0.258$^\diamond$&0.344$^\dagger$&0.319$^\dagger$ \\
SOTA1 & {0.395}$^\#$&{0.428}$^\#$&{0.193}$^\#$&{0.313}$^\#$&{0.367}$^\#$&{0.431}$^\#$&{0.452}$^\#$&{0.395}$^\#$&{0.333}$^\#$&\textbf{0.296}$^\#$&{0.255}$^\#$&{0.259}$^\#$&{0.381}$^\#$&{0.318}$^\#$&\textbf{0.377}$^\#$&{0.386}$^\#$ \\
{\ours }  & \textcolor{gg}{\textbf{0.431}} &\textcolor{gg}{\textbf{0.450}}&\textbf{0.197}&\textcolor{gg}{\textbf{0.411}}&\textcolor{gg}{\textbf{0.380}}&\textcolor{gg}{\textbf{0.448}}&\textcolor{gg}{\textbf{0.467}}&\textcolor{gg}{\textbf{0.452}}&\textcolor{gg}{\textbf{0.368}}&0.212&\textcolor{gg}{\textbf{0.309}}&\textcolor{gg}{\textbf{0.333}}&\textcolor{gg}{\textbf{0.404}}&\textcolor{gg}{\textbf{0.333}}&0.334&\textcolor{gg}{\textbf{0.408}} \\
\hline \hline
$\Delta_{SOTA}$	& \textcolor{gg}{$\uparrow $ {0.036}} & \textcolor{gg}{$\uparrow $ {0.022}} & \textcolor{gg}{$\uparrow $ {0.004}} & \textcolor{gg}{$\uparrow $ {0.097}} & \textcolor{gg}{$\uparrow $ {0.013}} & \textcolor{gg}{$\uparrow $ {0.017}} & \textcolor{gg}{$\uparrow $ {0.015}} & \textcolor{gg}{$\uparrow $ {0.057}} & \textcolor{gg}{$\uparrow $ {0.035}} & -- & \textcolor{gg}{$\uparrow $ {0.054}} & \textcolor{gg}{$\uparrow $ {0.074}} & \textcolor{gg}{$\uparrow $ {0.023}} & \textcolor{gg}{$\uparrow $ {0.015}} & -- & \textcolor{gg}{$\uparrow $ {0.022}} \\ 
\hline
\end{tabular}

\vspace{1mm}

\fontsize{7}{10}\selectfont
%\centering
\setlength\tabcolsep{0.7pt}
\begin{tabular}{|c || *{5}{K{1.09cm}} || *{2}{K{1.09cm}} || *{2}{K{1.09cm}} || *{2}{K{1.09cm}}|}
\hline
Dataset       & \multicolumn{5}{c||}{{CMU-MOSI}} & \multicolumn{2}{c||}{{ICT-MMMO}} & \multicolumn{2}{c||}{{YouTube}} & \multicolumn{2}{c|}{{MOUD}}  \\
Task			& \multicolumn{5}{c||}{Sentiment} & \multicolumn{2}{c||}{Sentiment} & \multicolumn{2}{c||}{Sentiment} & \multicolumn{2}{c|}{Sentiment} \\
Metric  & Acc\_$7$ & Acc\_$2$ & F1 & MAE & $r$ & Acc\_$2$ & F1 & Acc\_$3$ & F1 & Acc\_$2$ & F1 \\ 
\hline \hline
SOTA2 & 34.1$^\#$	& 77.1$^\ddagger$ & 77.0$^\ddagger$  & 0.968$^\ddagger$ & 0.625$^\ddagger$ & 72.5$^\ast$ & 72.6$^\ast$ & 48.3$^\ddagger$ & 45.1$^\dagger$ & {81.1}$^\#$ & 80.4$^\#$ \\ 
SOTA1 & 34.7$^\ddagger$	& {77.4}$^\#$   & {77.3}$^\#$  & 0.965$^\#$ & 0.632$^\#$ & 73.8$^\#$ & 73.1$^\#$ & 51.7$^\#$ & 51.6$^\#$ & {81.1}$^\ddagger$ & {81.2}$^\ddagger$ \\ 
{\ours} & \textcolor{gg}{\textbf{36.2}}   & \textbf{78.1} & \textbf{78.1} & \textbf{0.951}	& \textcolor{gg}{\textbf{0.662}} & \textcolor{gg}{\textbf{81.3}} & \textcolor{gg}{\textbf{79.2}} & \textcolor{gg}{\textbf{53.3}} & \textcolor{gg}{\textbf{52.4}} & \textbf{82.1} & \textbf{81.7} \\
%\Xhline{0.5\arrayrulewidth}
\hline \hline
$\Delta_{SOTA}$	& \textcolor{gg}{$\uparrow $ {1.5}} & \textcolor{gg}{$\uparrow $ {0.7}} & \textcolor{gg}{$\uparrow $ {0.8}} & \textcolor{gg}{$\downarrow $ {0.014}} & \textcolor{gg}{$\uparrow $ {0.030}} & \textcolor{gg}{$\uparrow $ {7.5}} & \textcolor{gg}{$\uparrow $ {6.1}} & \textcolor{gg}{$\uparrow $ {1.6}} & \textcolor{gg}{$\uparrow $ {0.8}} & \textcolor{gg}{$\uparrow $ {1.0}} & \textcolor{gg}{$\uparrow $ {0.5}}  \\ 
\hline
\end{tabular}

\vspace{1mm}

\fontsize{7}{10}\selectfont
%\centering
\setlength\tabcolsep{0.7pt}
\begin{tabular}{|c || *{12}{K{1.02cm}} |} % || *{4}{K{0.748cm}}|}
\hline
Dataset		& \multicolumn{12}{c|}{{IEMOCAP Emotions}} \\
Task		& \multicolumn{2}{c}{Happy} & \multicolumn{2}{c}{Sad} & \multicolumn{2}{c}{Angry} & \multicolumn{2}{c}{Frustrated} & \multicolumn{2}{c}{Excited} & \multicolumn{2}{c|}{Neutral} \\
Metric		& Acc\_$2$ & F1 & Acc\_$2$ & F1 & Acc\_$2$ & F1 & Acc\_$2$ & F1 & Acc\_$2$ & F1 & Acc\_$2$ & F1 \\
\hline \hline
SOTA2		& 86.7$^\ddagger$ & 84.2$^\S$ & 83.4$^\ast$ & 81.7$^\dagger$ & 85.1$^\diamond$ & 84.5$^\S$ & 79.5$^\ddagger$ & 76.6$^\ddagger$ & 89.6$^\ddagger$ & 86.3$^\#$ & 68.8$^\S$ & 67.1$^\S$ \\
SOTA1		& 90.1$^\#$ & 85.3$^\#$ & 85.8$^\#$ & 82.8$^\ast$ & 87.0$^\#$ & 86.0$^\#$ & 80.3$^\#$ & \textbf{76.8}$^\#$ & 89.8$^\#$ & \textbf{87.1}$^\ddagger$ & 71.8$^\#$ & \textbf{68.5}$^\S$ \\
{\ours}      & \textbf{90.2} & \textcolor{gg}{\textbf{85.8}} & \textcolor{gg}{\textbf{88.4}} & \textcolor{gg}{\textbf{86.1}} & \textcolor{gg}{\textbf{87.5}} & \textcolor{gg}{\textbf{86.7}} & \textbf{80.4} & 74.5 & \textbf{90.0} & \textbf{87.1} & \textbf{72.1} & 68.1 \\
\hline \hline
$\Delta_{SOTA}$	& \textcolor{gg}{$\uparrow $ {0.1}} & \textcolor{gg}{$\uparrow $ {0.5}} & \textcolor{gg}{$\uparrow $ {2.6}} & \textcolor{gg}{$\uparrow $ {3.3}} & \textcolor{gg}{$\uparrow $ {0.5}} & \textcolor{gg}{$\uparrow $ {0.7}} & \textcolor{gg}{$\uparrow $ {0.1}} & -- & \textcolor{gg}{$\uparrow $ {0.2}} & -- & \textcolor{gg}{$\uparrow $ {0.3}} & -- \\
\hline
\end{tabular}
\label{overall}
\vspace{-4mm}
\end{table*}

In this section, we transition to more challenging multimodal time series datasets. All the datasets consist of monologue videos. Features are extracted from the language (GloVe word embeddings~\citep{pennington2014glove}), visual (Facet~\citep{emotient}), and acoustic (COVAREP~\citep{degottex2014covarep}) modalities. For a detailed description of feature extraction, please refer to the appendix. 

We consider the following six datasets across three domains: {1) Multimodal Personality Trait Recognition}: \textbf{POM}~\citep{Park:2014:CAP:2663204.2663260} contains 903 movie review videos annotated for the following personality traits: confident (con), passionate (pas), voice pleasant (voi), dominant (dom), credible (cre), vivid (viv), expertise (exp), entertaining (ent), reserved (res), trusting (tru), relaxed (rel), outgoing (out), thorough (tho), nervous (ner), persuasive (per) and humorous (hum). The short form is indicated in parenthesis. {2) Multimodal Sentiment Analysis}: \textbf{CMU-MOSI}~\citep{zadeh2016multimodal} is a collection of 2199 monologue opinion video clips annotated with sentiment. \textbf{ICT-MMMO}~\citep{wollmer2013youtube} consists of 340 online social review videos annotated for sentiment. \textbf{YouTube}~\citep{morency2011towards} contains 269 product review and opinion video segments from YouTube each annotated for sentiment. \textbf{MOUD}~\citep{perez-rosas_utterance-level_2013} consists of 79 product review videos in Spanish. Each video consists of multiple segments labeled as either positive, negative or neutral sentiment. {3) Multimodal Emotion Recognition}: \textbf{IEMOCAP}~\citep{Busso2008IEMOCAP:Interactiveemotionaldyadic} consists of 302 videos of recorded dyadic dialogues. The videos are divided into multiple segments each annotated for the presence of 6 discrete emotions (happy, sad, angry, frustrated, excited and neutral), resulting in a total of 7318 segments in the dataset. We report results using the following metrics: Acc\_$C$ = multiclass accuracy across $C$ classes, F1 = F1 score, MAE = Mean Absolute Error, $r$ = Pearson's correlation.

\textbf{Prediction:} We first compare the performance of \ours \ with existing multimodal prediction methods. For a detailed description of the baselines, please refer to the appendix. From Table~\ref{overall}, we first observe that the best performing baseline results are achieved by different models across different datasets (most notably MFN, MARN, and TFN). On the other hand, \ours \ consistently achieves state-of-the-art or competitive results for all six multimodal datasets. We believe that the multimodal discriminative factor $\mathbf{F_y}$ in \ours \ has successfully learned more meaningful representations by distilling discriminative features. This highlights the benefit of learning factorized multimodal representations towards discriminative tasks. 
Furthermore, \ours \ is \textit{model-agnostic} and can be applied to other multimodal encoders $Q(\mathbf{Z_y}|\mathbf{X}_{1:M})$. We perform experiments to show consistent improvements in discriminative performance for several choices of the encoder: EF-LSTM~\citep{morency2011towards} and TFN~\citep{tensoremnlp17}. For Acc\_$2$ on CMU-MOSI, our factorization framework improves the performance of EF-LSTM from $74.3$ to $\textbf{75.2}$ and TFN from $74.6$ to $\textbf{75.5}$.

\begin{figure*}[t!]
\vspace{-3mm}
\centering
\fontsize{7}{10}\selectfont
%\centering
\setlength\tabcolsep{1.2pt}
\centering
\begin{tabular}{|*{1}{K{0.7cm}}||*{3}{K{1.2cm}:}*{1}{K{1.3cm}|}|*{3}{K{1.1cm}}||*{5}{K{0.7cm}}|}
\hline
\multirow{3}{*}{Model} & Multimodal               & Hybrid                   & Factorized               & Mod.-Spec. & \multicolumn{8}{c|}{CMU-MOSI}    \\ \cline{6-13} 
                          & Disc.                    & Gen.-Disc.               & Gen.-Disc.               & Gen.          & \multicolumn{3}{c||}{ $\hat{\mathbf{X}}_\cdot$ Reconstruction}  & \multicolumn{5}{c|}{ $\hat{\mathbf{Y}}$ Prediction} \\
                          & Factor                   & Objective                & Factors                  & Factors              & MSE ($\ell$)           & MSE ($a$)           & MSE ($v$) & Acc\_$7$ & Acc\_$2$  & F1    & MAE   & $r$ \\ \hline \hline
$\mathbf{M_A}$       & \multicolumn{1}{c:}{no}  & \multicolumn{1}{c:}{no}  & \multicolumn{1}{c:}{-}   & -   &  -  &    -   &  -   & 33.2 & 75.2  & 75.2  & 1.020    & 0.616        \\
$\mathbf{M_B}$       & \multicolumn{1}{c:}{yes} & \multicolumn{1}{c:}{no}  & \multicolumn{1}{c:}{-}   & -  &     -       &       -     &   - & 34.1   & 77.4 & 77.3   & 0.965 & 0.632     \\
$\mathbf{M_C}$       & \multicolumn{1}{c:}{no}  & \multicolumn{1}{c:}{yes} & \multicolumn{1}{c:}{no}  & -  & 0.0413 & 0.0509 & 0.0220 & 34.8 & 75.9 & 76.0 & 0.979   & 0.640       \\
$\mathbf{M_D}$       & \multicolumn{1}{c:}{yes} & \multicolumn{1}{c:}{yes} & \multicolumn{1}{c:}{no}  & -  & 0.0413 & 0.0486 & 0.0223 & 35.0& 77.4 & 77.2  & 0.960  & 0.649 \\
$\mathbf{M_E}$       & \multicolumn{1}{c:}{yes} & \multicolumn{1}{c:}{yes} & \multicolumn{1}{c:}{yes} & no  &    0.0397 &   0.0452   &  0.0211 & 35.9 & 77.3  & 77.2   & 0.956 & 0.661 \\ \hline \hline
MFM                  & \multicolumn{1}{c:}{yes} & \multicolumn{1}{c:}{yes} & \multicolumn{1}{c:}{yes} & yes  & \textbf{0.0391}  & \textbf{0.0384}  & \textbf{0.0183} & \textbf{36.2} & \textbf{78.1} & \textbf{78.1}  & \textbf{0.951} & \textbf{0.662}   \\ \hline
\end{tabular}
\includegraphics[width=1.0\textwidth]{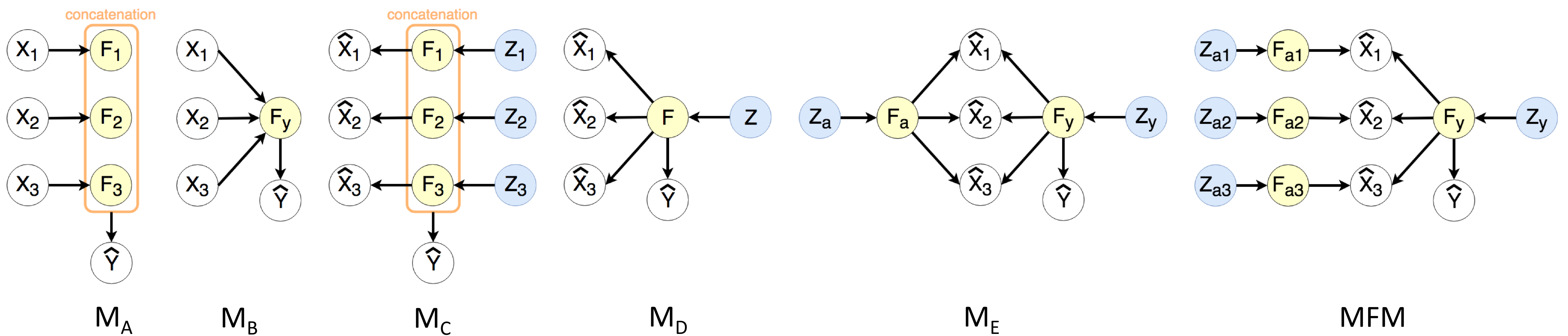}
\vspace{-5mm}
\caption{\small Models used in the ablation studies of \ours. Each model removes a design component from our model. Modality reconstruction and sentiment prediction results are reported on CMU-MOSI with best results in bold. Factorizing multimodal representations into multimodal discriminative factors and modality-specific generative factors are crucial for improved performance.}
\label{tbl:ablations}
\vspace{-8mm}
\end{figure*}

\textbf{Ablation Study:} In Figure~\ref{tbl:ablations}, we present the models $\mathbf{M_{\{A,B,C,D,E\}}}$ used for ablation studies. These models are designed to analyze the effects of using a multimodal discriminative factor, a hybrid generative-discriminative objective, factorized generative-discriminative factors and modality-specific generative factors towards both modality reconstruction and label prediction. The simplest variant is $\mathbf{M_{A}}$ which represents a purely discriminative model without a joint multimodal discriminative factor (i.e. early fusion~\citep{morency2011towards}). $\mathbf{M_{B}}$ models a joint multimodal discriminative factor which incorporates more general multimodal fusion encoders~\citep{zadeh2018memory}. $\mathbf{M_{C}}$ extends $\mathbf{M_A}$ by optimizing a hybrid generative-discriminative objective over modality-specific factors. $\mathbf{M_{D}}$ extends $\mathbf{M_B}$ by optimizing a hybrid generative-discriminative objective over a joint multimodal factor (resembling~\cite{srivastava2012multimodal}). $\mathbf{M_{E}}$ factorizes the representation into separate generative and discriminative factors. Finally, \ours \ is obtained from $\mathbf{M_E}$ by using modality-specific generative factors instead of a joint multimodal generative factor.

From the table in Figure~\ref{tbl:ablations}, we observe the following general trends. For sentiment prediction, using 1) a multimodal discriminative factor outperforms modality-specific discriminative factors ($\mathbf{M_D} > \mathbf{M_C}$, $\mathbf{M_B} > \mathbf{M_A}$), and 2) adding generative capabilities to the model improves performance ($\mathbf{M_C} > \mathbf{M_A}$, $\mathbf{M_E} > \mathbf{M_B}$). For both sentiment prediction and modality reconstruction, 3) factorizing into separate generative and discriminative factors improves performance ($\mathbf{M_E} > \mathbf{M_D}$), and 4) using modality-specific generative factors outperforms multimodal generative factors (\ours \ $ > \mathbf{M_E}$). These observations support our design decisions of factorizing multimodal representations into multimodal discriminative factors and modality-specific generative factors.

{\bf Missing Modalities:} We now evaluate the performance of \ours \ in the presence of missing modalities using the surrogate inference model as described in Subsection~\ref{sec:missing}. We compare with two baselines: 1) a purely generative Seq2Seq model~\citep{cho-al-emnlp14} $\Phi_G$ from observed modalities to missing modalities by optimizing $\mathbf{E}_{P_{\mathbf{X}_{1:M}}}\, (-\mathrm{log}\,P_{\Phi_D}(\mathbf{X}_1|\mathbf{X}_{2:M}))$, and 2) a purely discriminative model $\Phi_D$ from observed modalities to the label by optimizing $\mathbf{E}_{P_{\mathbf{X}_{2:M}, \mathbf{Y}}}\, ( -\mathrm{log}\,P_{\Phi_G}(\mathbf{Y}|\mathbf{X}_{2:M}))$. Both models are modified from \ours \ by using only the two observed modalities as input and not explicitly accounting for missing modalities. We compare the reconstruction error of each modality (language, visual and acoustic) as well as the performance on sentiment prediction.

Table~\ref{tbl:missing} shows that \ours \ with missing modalities outperforms the generative ($\Phi_G$) or discriminative baselines ($\Phi_D$) in terms of modality reconstruction and sentiment prediction. Additionally, \ours \ with missing modalities performs close to \ours \ with all modalities observed. This fact indicates that \ours \ can learn representations that are relatively robust to missing modalities. In addition, discriminative performance is most affected when the language modality is missing, which is consistent with prior work which indicates that language is most informative in human multimodal language~\citep{tensoremnlp17}. On the other hand, sentiment prediction is more robust to missing acoustic and visual features. Finally, we observe that reconstructing the low-level acoustic and visual features is easier as compared to the high-dimensional language features that contain high-level semantic meaning.

\begin{center}
\begin{table}[t!]
\vspace{-5mm}
\caption{\small The effect of missing modalities on multimodal data reconstruction and sentiment prediction on CMU-MOSI. \ours \ with surrogate inference is able to better handle missing modalities during test time as compared to the purely generative (Seq2Seq) or purely discriminative baselines.}
\fontsize{7}{10}\selectfont
\centering
\setlength\tabcolsep{5pt}
\vspace{-3mm}
\begin{tabular}{|c||ccc||ccccc|}
\hline
Task        & \multicolumn{3}{c||}{$\hat{\mathbf{X}}_\cdot$ Reconstruction} & \multicolumn{5}{c|}{$\hat{\mathbf{Y}}$ Prediction} \\
Metric      & MSE ($\ell$)   & MSE ($a$)   & MSE ($v$) & Acc\_$7$  & Acc\_$2$  & F1   & MAE & $r$ \\ \hline \hline
\multicolumn{9}{|c|}{Purely Generative and Discriminative Baselines}  \\ \hline \hline
$\ell$(anguage) missing  & 0.0411 & - & - & 19.4 & 59.6 & 59.7 & 1.386 & 0.225 \\
$a$(udio) missing  & - & 0.0533 & - & 34.0& 73.5 & 73.4  & 1.024 & 0.615 \\
$v$(isual) missing  & - & - & 0.0220 & 33.7 & 75.4 & 75.4  & 0.996 & 0.634 \\ \hline \hline
\multicolumn{9}{|c|}{\ourl \ (\ours) } \\ \hline \hline
$\ell$(anguage) missing   & 0.0403 & - & - & 21.7 & 62.0 & 61.7  & 1.313 & 0.236 \\
$a$(udio) missing   & - & 0.0468 & - &  35.4  &  74.3 & 74.3 & 1.011 & 0.603 \\
$v$(isual) missing  & - & - & 0.0215 &  35.0 & 76.4  & 76.3  & 0.990 & 0.635 \\ 
all present & \textbf{0.0391}  & \textbf{0.0384}  & \textbf{0.0182} & \textbf{36.2} &  \textbf{78.1} & \textbf{78.1}  & \textbf{0.951} & \textbf{0.662}  \\ \hline
\end{tabular}
\vspace{-4mm}
\label{tbl:missing}
\end{table}
\end{center}

\vspace{-9mm}
{\bf Interpretation of Multimodal Representations:} We devise two methods to study how individual factors in \ours \ influence the dynamics of multimodal prediction and generation. These interpretation methods represent both overall trends and fine-grained analysis that could be useful towards deeper understandings of multimodal representation learning. For more details, please refer to the appendix.

\begin{figure}[t!]
\vspace{-5mm}
  \begin{minipage}[c]{0.42\textwidth}
    	\fontsize{7}{10}\selectfont
	\centering
	\vspace{-4mm}
    \begin{tabular}{| c || *{3}{K{0.8cm}}|}
\hline
Ratio		& $r_\ell$ & $r_v$ & $r_a$ \\ 
\hline \hline
CMU-MOSI	& 0.307 & 0.030 & 0.107 \\
\hline
\end{tabular}
    \captionsetup{labelformat=simple}
	\caption{\small Analyzing the multimodal representations learnt in \ours \ via information-based (entire dataset) and gradient-based interpretation methods (single video) on CMU-MOSI.
    \vspace{-5mm}}
	\label{fig:interp}
  \end{minipage}\hfill
  \begin{minipage}[c]{0.5\textwidth}
    \includegraphics[width=\linewidth]{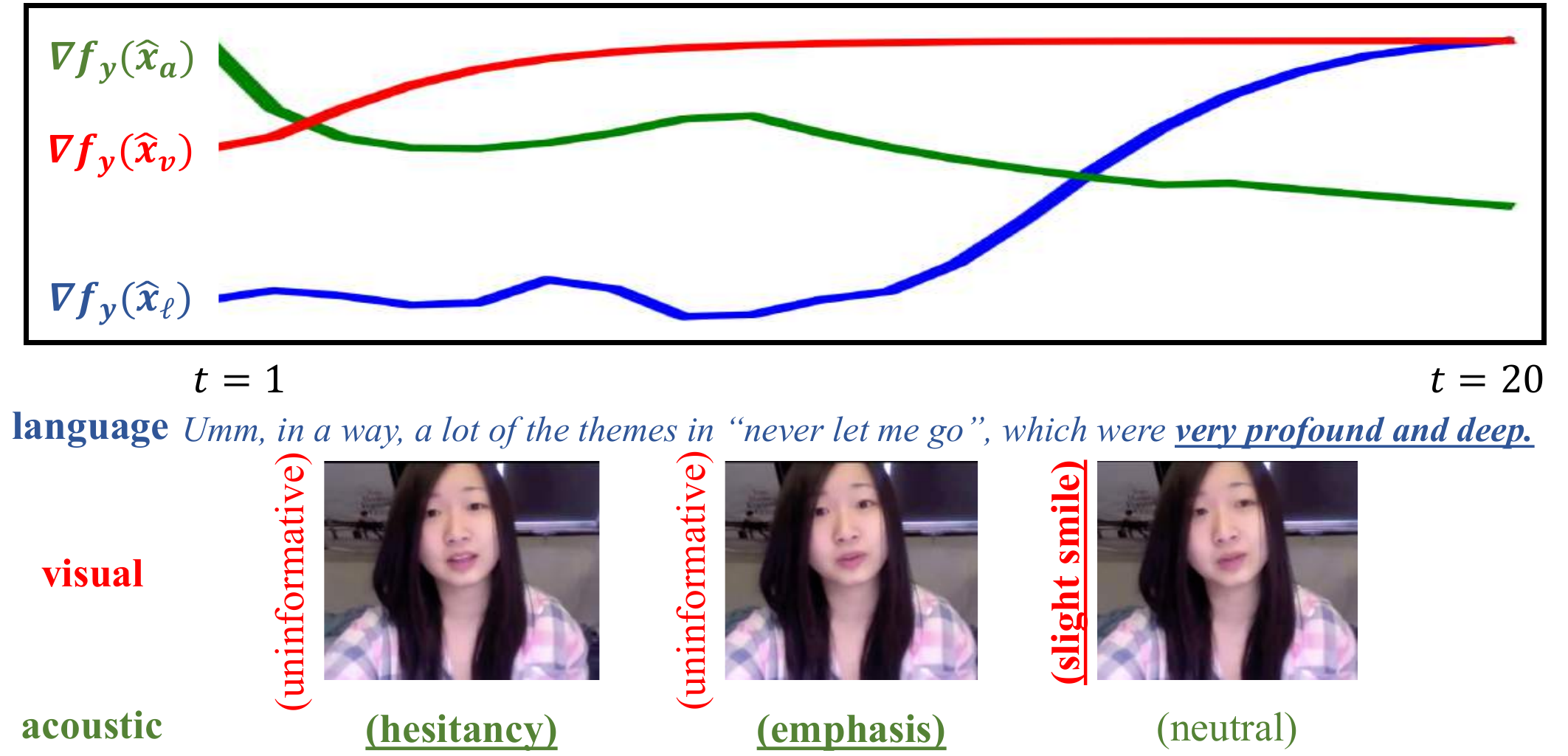}
  \end{minipage}
  \vspace{-4mm}
\end{figure}

Firstly, an information-based interpretation method is chosen to summarize the contribution of each modality towards the multimodal representations. Since $\mathbf{F_y}$ is a common cause of $\hat{\mathbf{X}}_{1:M}$, we can compare $\textnormal{MI}(\mathbf{F_y}, \hat{\mathbf{X}}_1),$ $\cdots, \textnormal{MI}(\mathbf{F_y}, \hat{\mathbf{X}}_M)$, where $\textnormal{MI}(\cdot, \cdot)$ denotes the mutual information measure between $\mathbf{F_y}$ and generated modality $\hat{\mathbf{X}}_i$. Higher $\textnormal{MI}(\mathbf{F_y}, \hat{\mathbf{X}}_i)$ indicates greater contribution from $\mathbf{F_y}$ to $\hat{\mathbf{X}}_i$. Figure~\ref{fig:interp} reports the ratios $r_i = \textrm{MI}(\mathbf{F_y},\hat{\mathbf{X}}_{i}) / \textrm{MI}(\mathbf{F_a}_{i},\hat{\mathbf{X}}_{i})$ which measure a normalized version of the mutual information between $\mathbf{F_a}_{i}$ and $\hat{\mathbf{X}}_{i}$. We observe that on CMU-MOSI, the language modality is most informative towards sentiment prediction, followed by the acoustic modality. We believe that this result represents a prior over the expression of sentiment in human multimodal language and is closely related to the connections between language and speech \citep{Kuhl11850}. 

Secondly, a gradient-based interpretation method to used analyze the contribution of each modality for every time step in multimodal time series data. We measure the gradient of the generated modality with respect to the target factors (e.g., $\mathbf{F_y}$). Let $\{x_1, x_2, \cdots, x_M\}$ denote multimodal time series data where $x_i$ represents modality $i$, and $\hat{x}_i = [\hat{x}_i^1, \cdots, \hat{x}_i^t, \cdots, \hat{x}_i^T]$ denote generated modality $i$ across time steps $t \in [1,T]$. The gradient $\nabla_{f_{y}} (\hat{x}_i)$ measures the extent to which changes in factor $f_y \sim P(\mathbf{F_y}|\mathbf{X}_{1:M} = x_{1:M})$ influences the generation of sequence $\hat{x}_i$. Figure~\ref{fig:interp} plots $\nabla_{f_{y}} (\hat{x}_i)$ for a video in CMU-MOSI. We observe that multimodal communicative behaviors that are indicative of speaker sentiment such as positive words (e.g. ``very profound and deep'') and informative acoustic features (e.g. hesitant and emphasized tone of voice) indeed correspond to increases in $\nabla_{f_{y}} (\hat{x}_i)$. 

\vspace{-2mm}
\section{Related Work}

The two main pillars of research in multimodal representation learning have considered the discriminative and generative objectives individually. Discriminative representation learning~\citep{localglobal,Chen:2017:MSA:3136755.3136801,chaplot2017gated,frome2013devise,socher2013zero,tsai2017learning} models the conditional distribution $P(\mathbf{Y}|\mathbf{X}_{1:M})$. Since these approaches are not concerned with modeling $P(\mathbf{X}_{1:M})$ explicitly, they use parameters more efficiently to model $P(\mathbf{Y}|\mathbf{X}_{1:M})$. For instance, recent works learn visual representations that are maximally dependent with linguistic attributes for improving one-shot image recognition~\citep{tsai2017improving} or introduce tensor product mechanisms to model interactions between the language, visual and acoustic modalities~\citep{lowrank,tensoremnlp17}. On the other hand, generative representation learning captures the interactions between modalities by modeling the joint distribution $P(\mathbf{X}_1, \cdots , \mathbf{X}_M)$ using either undirected graphical models~\citep{srivastava2012multimodal}, directed graphical models~\citep{suzuki2016joint}, or neural networks~\citep{sohn2014improved}. Some generative approaches compress multimodal data into lower-dimensional feature vectors which can be used for discriminative tasks~\citep{seq2seq,ngiam2011multimodal}. To unify the advantages of both approaches, \ours \ factorizes multimodal representations into generative and discriminative components and optimizes for a joint objective. 

Factorized representation learning resembles learning disentangled data representations which have been shown to improve the performance on many tasks~\citep{kulkarni2015deep,lake2017building,higgins2016beta,Bengio:2013:RLR:2498740.2498889}. Several methods involve specifying a fixed set of latent attributes that individually control particular variations of data and performing supervised training~\citep{cheung2014discovering,karaletsos2015bayesian,yang2015weakly,reed2014learning,zhu2014multi}, assuming an isotropic Gaussian prior over latent variables to learn disentangled generative representations~\citep{kingma2013auto,rubenstein2018latent} and learning latent variables in charge of specific variations in the data by maximizing the mutual information between a subset of latent variables and the data~\citep{chen2016infogan}. However, these methods study factorization of a single modality. \ours \ factorizes multimodal representations and demonstrates the importance of modality-specific and multimodal factors towards generation and prediction. A concurrent and parallel work that factorizes latent factors in multimodal data was proposed by~\cite{hsu2018disentangling}. They differ from us in the graphical model design, discriminative objective, prior matching criterion, and scale of experiments. We provide a detailed comparison with their model in the appendix.
\vspace{-2mm}
\section{Conclusion}

In this paper, we proposed the \ourl \ (\ours) for multimodal representation learning. \ours \ factorizes the multimodal representations into two sets of independent factors: \textit{multimodal discriminative} factors and \textit{modality-specific generative} factors. The multimodal discriminative factor achieves state-of-the-art or competitive results on six multimodal datasets. The modality-specific generative factors allow us to generate data based on factorized variables, account for missing modalities, and have a deeper understanding of the interactions involved in multimodal learning. Our future work will explore extensions of \ours \ for video generation, semi-supervised learning, and unsupervised learning. We believe that \ours \ sheds light on the advantages of learning factorizing multimodal representations and potentially opens up new horizons for multimodal machine learning.

\section*{Acknowledgements}
This work was supported in part by the DARPA grants D17AP00001 and FA875018C0150, Office of Naval Research, Apple, and Google focused award. We would also like to acknowledge NVIDIA's GPU support. This material is also based upon work partially supported by the National Science Foundation (Award \#1750439). Any opinions, findings, and conclusions or recommendations expressed in this material are those of the author(s) and do not necessarily reflect the views of National Science Foundation, and no official endorsement should be inferred.

{
\bibliography{iclr2019}
\bibliographystyle{iclr2019_conference}
}

\renewcommand{\thesection}{\Alph{section}}
\setcounter{section}{0}

\section{Proof of Proposition 1}

To simplify the proof, we first prove it for the unimodal case by considering the Wasserstein distance between ${P}_{\mathbf{X},\mathbf{Y}}$ and ${P}_{\mathbf{\hat{X}},\mathbf{\hat{Y}}}$.

\subsection{Unimodal Joint-Distribution Wasserstein Distance}

%!TEX root = ../nips_2018.tex

\begin{proposition}
\label{theo:wass}
For any functions $G_y: \mathbf{Z_y} \rightarrow \mathbf{F_y}$, $G_a: \mathbf{Z_a} \rightarrow \mathbf{F_a}$, $D: \mathbf{F_y} \rightarrow \mathbf{\hat{Y}}$, and $F: \mathbf{F_a, F_y} \rightarrow \mathbf{\hat{X}}$, we have 
{
\begin{equation}
\label{eq:theowass}
\begin{split}
W_c({P}_{\mathbf{X,Y}}, {P}_{\mathbf{\hat{X},\hat{Y}}}) =  \underset{{Q_{\mathbf{Z}}} = {P_{\mathbf{Z}}}}{\mathrm{inf}}\,\,\mathbf{E}_{{P}_{\mathbf{X,Y}}}\mathbf{E}_{{Q}({\mathbf{Z}|\mathbf{X}})}\Bigg[c_X\Big(\mathbf{X}, F\big(G_a(\mathbf{Z_a}), G_y(\mathbf{Z_y})\big)\Big) + c_Y\Big(\mathbf{Y}, D\big(G_y(\mathbf{Z_y})\big)\Big)\Bigg],
\end{split}
\end{equation}
} where $W_c$ is the Wasserstein distance under cost function $c_X$ and $c_Y$, ${P_{\mathbf{Z}}}$ is the prior over $\mathbf{Z} = [\mathbf{Z_a}, \mathbf{Z_y} ]$ and ${Q_{\mathbf{Z}}}$ is the aggregated posterior of the proposed inference distribution ${Q}({\mathbf{Z}|\mathbf{X}})$.
\end{proposition}

{\em Proof:} See the following.

To begin the proof, we abuse some notations as follows. 

By definition, the Wasserstein distance under cost function $c$ between ${P}_{\mathbf{X,Y}}$ and ${P}_{\mathbf{\hat{X},\hat{Y}}}$ is 
\begin{equation}
\label{eq:joint1}
\small
\begin{split}
&W_c({P}_{\mathbf{X,Y}}, {P}_{\mathbf{\hat{X},\hat{Y}}}) := \underset{ \Gamma\in \mathcal{P}\Big((\mathbf{X}, \mathbf{Y})\sim {P}_{\mathbf{X}, \mathbf{Y}}, (\mathbf{\hat{X}}, \mathbf{\hat{Y}})\sim {P}_{\mathbf{\hat{X}}, \mathbf{\hat{Y}}}\Big)}{\mathrm{inf}}\,\,\mathbf{E}_{\big((\mathbf{X}, \mathbf{Y}), (\mathbf{\hat{X}}, \mathbf{\hat{Y}})\big)\sim \Gamma } \Big[ c\Big((\mathbf{X}, \mathbf{Y}), (\mathbf{\hat{X}}, \mathbf{\hat{Y}})\Big)\Big],
\end{split}
\end{equation}
where $c\Big((\mathbf{X}, \mathbf{Y}), (\mathbf{\hat{X}}, \mathbf{\hat{Y}})\Big): (\mathcal{X}, \mathcal{Y}) \times (\mathcal{X}, \mathcal{Y}) \rightarrow \mathcal{R}_{+}$ is any measurable {\em cost function}. $\mathcal{P}\Big((\mathbf{X}, \mathbf{Y})\sim {P}_{\mathbf{X}, \mathbf{Y}}, (\mathbf{\hat{X}}, \mathbf{\hat{Y}})\sim {P}_{\mathbf{\hat{X}}, \mathbf{\hat{Y}}}\Big)$ is the set of all joint distributions of $\Big((\mathbf{X}, \mathbf{Y}), (\mathbf{\hat{X}}, \mathbf{\hat{Y}})\Big)$ with marginals ${P}_{\mathbf{X,Y}}$ and ${P}_{\mathbf{\hat{X},\hat{Y}}}$, respectively. Note that $c\Big((\mathbf{X}, \mathbf{Y}), (\mathbf{\hat{X}}, \mathbf{\hat{Y}})\Big) = c_{X}\Big(\mathbf{X}, \mathbf{\hat{X}}\Big) + c_{Y}\Big(\mathbf{Y}, \mathbf{\hat{Y}}\Big)$. 

Next, we denote the set of all joint distributions of ($\mathbf{X}, \mathbf{Y}, \mathbf{\hat{X}}, \mathbf{\hat{Y}}, \mathbf{Z}$) such that $(\mathbf{X}, \mathbf{Y})\sim {P}_{\mathbf{X}, \mathbf{Y}}$, $(\mathbf{\hat{X}}, \mathbf{\hat{Y}}, \mathbf{Z})\sim {P}_{\mathbf{\hat{X}}, \mathbf{\hat{Y}}, \mathbf{Z}}$, and $\Big((\mathbf{X}, \mathbf{Y}) \independent (\mathbf{\hat{X}}, \mathbf{\hat{Y}}) | \mathbf{Z}\Big)$ as $\mathcal{P}_{\mathbf{X}, \mathbf{Y}, \mathbf{\hat{X}}, \mathbf{\hat{Y}}, \mathbf{Z}}$. $\mathcal{P}_{\mathbf{X}, \mathbf{Y}, \mathbf{\hat{X}}, \mathbf{\hat{Y}}}$ and $\mathcal{P}_{\mathbf{X}, \mathbf{Y}, \mathbf{Z}}$ are the sets of the marginals $(\mathbf{X, Y, \hat{X}, \hat{Y}})$  and $(\mathbf{X, Y, Z})$ induced by $\mathcal{P}_{\mathbf{X}, \mathbf{Y}, \mathbf{\hat{X}}, \mathbf{\hat{Y}}, \mathbf{Z}}$.

We now introduce two Lemmas to help the proof.
\begin{lemma}
\label{lem:1}
$P(\mathbf{\hat{X}}, \mathbf{\hat{Y}}|\mathbf{Z} = z)$ are Dirac for all $z \in \mathcal{Z}$.
\end{lemma}
{\em Proof:} 
First, we have $\mathbf{\hat{X}} = F(G_a(\mathbf{Z_a}), G_y(\mathbf{Z_y}))$ and $\mathbf{\hat{Y}} = D(G_y(\mathbf{Z_y}))$ with $\mathbf{Z} = \{\mathbf{Z_a}, \mathbf{Z_y}\}$. Since the functions $F, G_a, G_y, D$ are all deterministic, then $P(\mathbf{\hat{X}}, \mathbf{\hat{Y}}|\mathbf{Z})$ are Dirac measures.
\QEDB

\begin{lemma}
\label{lem:2}
 $\mathcal{P}\Big({P}_{\mathbf{X}, \mathbf{Y}}, {P}_{\mathbf{\hat{X}}, \mathbf{\hat{Y}}}\Big)$ = $\mathcal{P}_{\mathbf{X}, \mathbf{Y}, \mathbf{\hat{X}}, \mathbf{\hat{Y}}}$ when $P(\mathbf{\hat{X}}, \mathbf{\hat{Y}}|\mathbf{Z} = z)$ are Dirac for all $z \in \mathcal{Z}$.
\end{lemma}
{\em Proof:} 
When $\mathbf{\hat{X}}, \mathbf{\hat{Y}}$ are deterministic functions of $\mathbf{Z}$, for any $A$ in the sigma-algebra induced by $\mathbf{\hat{X}}, \mathbf{\hat{Y}}$, we have 
$$
\mathbf{E}[\mathbb{I}_{[\mathbf{\hat{X}}, \mathbf{\hat{Y}} \in A]}|\mathbf{X}, \mathbf{Y}, \mathbf{Z}] = \mathbf{E}[\mathbb{I}_{[\mathbf{\hat{X}}, \mathbf{\hat{Y}} \in A]}|\mathbf{Z}].
$$
Therefore, this implies that $(\mathbf{X}, \mathbf{Y}) \independent (\mathbf{\hat{X}}, \mathbf{\hat{Y}}) | \mathbf{Z}$ which concludes the proof. A similar argument is made in Lemma 1 of~\citep{tolstikhin2017wasserstein}.

\QEDB

Now, we use the fact that $\mathcal{P}\Big({P}_{\mathbf{X}, \mathbf{Y}}, {P}_{\mathbf{\hat{X}}, \mathbf{\hat{Y}}}\Big)$ = $\mathcal{P}_{\mathbf{X}, \mathbf{Y}, \mathbf{\hat{X}}, \mathbf{\hat{Y}}}$ (Lemma \ref{lem:1} + Lemma \ref{lem:2}), $c\Big((\mathbf{X}, \mathbf{Y}), (\mathbf{\hat{X}}, \mathbf{\hat{Y}})\Big) = c_{X}\Big(\mathbf{X}, \mathbf{\hat{X}}\Big) + c_{Y}\Big(\mathbf{Y}, \mathbf{\hat{Y}}\Big)$, $\mathbf{\hat{X}} = F\big(G_a(\mathbf{Z_a}), G_y(\mathbf{Z_y})\big)$, and $\mathbf{\hat{Y}} = D\big(G_y(\mathbf{Z_y})\big)$, Eq. \eqref{eq:joint1} becomes
\begin{equation}
\label{eq:joint3}
\begin{split}
&\underset{ P\in \mathcal{P}_{\mathbf{X}, \mathbf{Y}, \mathbf{\hat{X}}, \mathbf{\hat{Y}}}}{\mathrm{inf}}\,\,\mathbf{E}_{\mathbf{X}, \mathbf{Y}, \mathbf{\hat{X}}, \mathbf{\hat{Y}}\sim P } \Big[ c_X\Big(\mathbf{X}, \mathbf{\hat{X}}\Big) + c_Y\Big(\mathbf{Y}, \mathbf{\hat{Y}}\Big)\Big]\\
=&\underset{ P\in \mathcal{P}_{\mathbf{X}, \mathbf{Y}, \mathbf{\hat{X}}, \mathbf{\hat{Y}}, \mathbf{Z}}}{\mathrm{inf}}\,\,\mathbf{E}_{\mathbf{X}, \mathbf{Y}, \mathbf{\hat{X}}, \mathbf{\hat{Y}}, \mathbf{Z} \sim P } \Big[ c_X\Big(\mathbf{X}, \mathbf{\hat{X}}\Big) + c_Y\Big(\mathbf{Y}, \mathbf{\hat{Y}}\Big)\Big]\\
=&\underset{ P\in \mathcal{P}_{\mathbf{X}, \mathbf{Y}, \mathbf{\hat{X}}, \mathbf{\hat{Y}}, \mathbf{Z}}}{\mathrm{inf}}\,\,\mathbf{E}_{{P}_{\mathbf{Z}}}\mathbf{E}_{{P}(\mathbf{X,Y}|\mathbf{Z})}\mathbf{E}_{{P}(\mathbf{\hat{X},\hat{Y}}|\mathbf{Z})}\Big[ c_X\Big(\mathbf{X}, \mathbf{\hat{X}}\Big) + c_Y\Big(\mathbf{Y}, \mathbf{\hat{Y}}\Big)\Big]\\
=&\underset{ P\in \mathcal{P}_{\mathbf{X}, \mathbf{Y}, \mathbf{\hat{X}}, \mathbf{\hat{Y}}, \mathbf{Z}}}{\mathrm{inf}}\,\,\mathbf{E}_{{P}_{\mathbf{Z}}}\mathbf{E}_{{P}(\mathbf{X,Y}|\mathbf{Z})}\Big[ c_X\Big(\mathbf{X}, F\big(G_a(\mathbf{Z_a}), G_y(\mathbf{Z_y})\big)\Big) + c_Y\Big(\mathbf{Y}, D\big(G_y(\mathbf{Z_y})\big)\Big)\Big]\\
=&\underset{ P\in \mathcal{P}_{\mathbf{X}, \mathbf{Y}, \mathbf{Z}}}{\mathrm{inf}}\,\,\mathbf{E}_{{P}_{\mathbf{Z}}}\mathbf{E}_{{P}(\mathbf{X,Y}|\mathbf{Z})}\Big[ c_X\Big(\mathbf{X}, F\big(G_a(\mathbf{Z_a}), G_y(\mathbf{Z_y})\big)\Big) + c_Y\Big(\mathbf{Y}, D\big(G_y(\mathbf{Z_y})\big)\Big)\Big]\\
=&\underset{ P\in \mathcal{P}_{\mathbf{X}, \mathbf{Y}, \mathbf{Z}}}{\mathrm{inf}}\,\,\mathbf{E}_{\mathbf{X,Y, Z}\sim P}\Big[ c_X\Big(\mathbf{X}, F\big(G_a(\mathbf{Z_a}), G_y(\mathbf{Z_y})\big)\Big) + c_Y\Big(\mathbf{Y}, D\big(G_y(\mathbf{Z_y})\big)\Big)\Big].
\end{split}
\end{equation}

Note that in Eq. \eqref{eq:joint3}, $\mathcal{P}_{\mathbf{X}, \mathbf{Y}, \mathbf{Z}} = \mathcal{P}\Big((\mathbf{X}, \mathbf{Y})\sim {P}_{\mathbf{X}, \mathbf{Y}}, \mathbf{Z}\sim {P}_{\mathbf{Z}}\Big)$ and with a proposed ${Q}(\mathbf{Z|X})$, we can rewrite Eq. \eqref{eq:joint3} as 
\begin{equation}
\label{eq:joint4}
\begin{split}
&\underset{ P\in \mathcal{P}_{\mathbf{X}, \mathbf{Y}, \mathbf{Z}}}{\mathrm{inf}}\,\,\mathbf{E}_{{P}_{\mathbf{X,Y}}}\mathbf{E}_{{P}_{\mathbf{Z}}}
\Big[ c_X\Big(\mathbf{X}, F\big(G_a(\mathbf{Z_a}), G_y(\mathbf{Z_y})\big)\Big) + c_Y\Big(\mathbf{Y}, D\big(G_y(\mathbf{Z_y})\big)\Big)\Big]\\
=&\underset{{Q_{\mathbf{Z}}} = {P_{\mathbf{Z}}}}{\mathrm{inf}}\,\,\mathbf{E}_{{P}_{\mathbf{X,Y}}}\mathbf{E}_{{Q}({\mathbf{Z}|\mathbf{X}})}\Bigg[c_X\Big(\mathbf{X}, F\big(G_a(\mathbf{Z_a}), G_y(\mathbf{Z_y})\big)\Big) + c_Y\Big(\mathbf{Y}, D\big(G_y(\mathbf{Z_y})\big)\Big)\Bigg].
\end{split}
\end{equation}

\QEDA

\subsection{From Unimodal to Multimodal}

The proof is similar to Proposition \ref{theo:wass}, and we present a sketch to it. We can first show $P(\mathbf{\hat{X}}_{1:M}, \mathbf{\hat{Y}}|\mathbf{Z} = z)$ are Dirac for all $z \in \mathcal{Z}$. Then we use the fact that $c\Big((\mathbf{X}_{1:M}, \mathbf{Y}), (\mathbf{\hat{X}}_{1:M}, \mathbf{\hat{Y}})\Big) = \sum_{i=1}^M {c_{X}}_i \Big(\mathbf{X}_i, \mathbf{\hat{X}}_i\Big) + c_{Y}\Big(\mathbf{Y}, \mathbf{\hat{Y}}\Big)$. Finally, we follow the tower rule of expectation and the conditional independence property similar to the proof in Proposition \ref{theo:wass} and this concludes the proof.

\QEDA

\section{Full Baseline Models \& Results}

For a detailed description of the baselines, we point the reader to MFN~\citep{zadeh2018memory}, MARN~\citep{zadeh2018multi}, TFN~\citep{tensoremnlp17}, {BC-LSTM}~\citep{contextmultimodalacl2017}, {MV-LSTM}~\citep{rajagopalan2016extending}, EF-LSTM~\citep{hochreiter1997long,6638947,Schuster:1997:BRN:2198065.2205129}, {DF}~\citep{Nojavanasghari:2016:DMF:2993148.2993176}, {MV-HCRF}~\citep{song2012multi,song2013action}, {EF-HCRF}~\citep{Quattoni:2007:HCR:1313053.1313265,morency2007latent}, {THMM}~\citep{morency2011towards}, {SVM-MD}~\citep{zadeh2016multimodal} and {RF}~\citep{Breiman:2001:RF:570181.570182}.

We use the following extra notations for full descriptions of the baseline models described in Section~\ref{disc}, paragraph 3:

Variants of EF-LSTM: \textbf{EF-LSTM} (Early Fusion LSTM) uses a single LSTM~\citep{hochreiter1997long} on concatenated multimodal inputs. We also implement the \textbf{EF-SLSTM} (stacked)~\citep{6638947}, \textbf{EF-BLSTM} (bidirectional)~\citep{Schuster:1997:BRN:2198065.2205129} and \textbf{EF-SBLSTM} (stacked bidirectional) versions.

Variants of EF-HCRF: \textbf{EF-HCRF}: (Hidden Conditional Random Field) \citep{Quattoni:2007:HCR:1313053.1313265} uses a HCRF to learn a set of latent variables conditioned on the concatenated input at each time step. \textbf{EF-LDHCRF} (Latent Discriminative HCRFs) \citep{morency2007latent} are a class of models that learn hidden states in a CRF using a latent code between observed concatenated input and hidden output. \textbf{EF-HSSHCRF}: (Hierarchical Sequence Summarization HCRF) \citep{song2013action} is a layered model that uses HCRFs with latent variables to learn hidden spatio-temporal dynamics.

Variants of MV-HCRF: \textbf{MV-HCRF}: Multi-view HCRF \citep{song2012multi} is an extension of the HCRF for Multi-view data, explicitly capturing view-shared and view specific sub-structures. \textbf{MV-LDHCRF}: \citep{morency2007latent} is a variation of the MV-HCRF model that uses LDHCRF instead of HCRF. \textbf{MV-HSSHCRF}: \citep{song2013action} further extends \textbf{EF-HSSHCRF} by performing Multi-view hierarchical sequence summary representation.

%\section{Full Results}

In the following, we provide the full results for all baselines models described in Section~\ref{disc}, paragraph 3. Table~\ref{pom} contains results for multimodal speaker traits recognition on the POM dataset. Table~\ref{mosi} contains results for the multimodal sentiment analysis on the CMU-MOSI, ICT-MMMO, YouTube, and MOUD datasets. Table~\ref{iemocap} contains results for multimodal emotion recognition on the IEMOCAP dataset. \ours \ consistently achieves state-of-the-art or competitive results for all six multimodal datasets. We believe that by our \ours \ design, the multimodal discriminative factor $\mathbf{F_y}$ has successfully learned more meaningful representations by distilling discriminative features. This highlights the benefit of learning factorized multimodal representations towards discriminative tasks. 
\section{Multimodal Features}

For each of the multimodal time series datasets as mentioned in Section~\ref{disc}, paragraph 3, we extracted the following multimodal features: \textbf{Language:} We use pre-trained word embeddings (glove.840B.300d)~\citep{pennington2014glove} to convert the video transcripts into a sequence of 300 dimensional word vectors. \textbf{Visual:} We use Facet~\citep{emotient} to extract a set of features including per-frame basic and advanced emotions and facial action units as indicators of facial muscle movement~\citep{ekman1980facial,ekman1992argument}. \textbf{Acoustic:} We use COVAREP~\citep{degottex2014covarep} to extract low level acoustic features including 12 Mel-frequency cepstral coefficients (MFCCs), pitch tracking and voiced/unvoiced segmenting features, glottal source parameters, peak slope parameters and maxima dispersion quotients. To reach the same time alignment between different modalities we choose the granularity of the input to be at the level of words. The words are aligned with audio using P2FA~\citep{P2FA} to get their exact utterance times. We use expected feature values across the entire word for visual and acoustic features since they are extracted at a higher frequencies. 

We make a note that the features for some of these datasets are constantly being updated. The authors of~\cite{zadeh2018memory} notified us of a discrepancy in the sampling rate for acoustic feature extraction in the ICT-MMMO, YouTube and MOUD datasets which led to inaccurate word-level alignment between the three modalities. They publicly released the updated multimodal features. We performed all experiments on the latest versions of these datasets which can be accessed from \url{https://github.com/A2Zadeh/CMU-MultimodalSDK}. All baseline models were retrained with extensive hyperparameter search for fair comparison.

\newcolumntype{K}[1]{>{\centering\arraybackslash}p{#1}}
%\definecolor{gg}{RGB}{45,190,45}
\definecolor{gg}{RGB}{0,0,0}

\begin{table*}[!htbp]
\fontsize{7}{10}\selectfont
%\centering
\caption{Results for personality trait recognition on the POM dataset. The best results are highlighted in bold and $\Delta_{SOTA}$ shows the change in performance over previous state of the art. Improvements are highlighted in green. \ours \ achieves state-of-the-art or competitive performance on all datasets and metrics.}
\setlength\tabcolsep{0.5pt}
\begin{tabular}{l : *{16}{K{0.75cm}}}
\Xhline{3\arrayrulewidth}
Dataset & \multicolumn{16}{c}{\textbf{POM Speaker Personality Traits}} \\
Task & \multicolumn{1}{c}{Con} & \multicolumn{1}{c}{Pas} & \multicolumn{1}{c}{Voi} & \multicolumn{1}{c}{Dom} & \multicolumn{1}{c}{Cre} & \multicolumn{1}{c}{Viv} & \multicolumn{1}{c}{Exp} & \multicolumn{1}{c}{Ent} & \multicolumn{1}{c}{Res} & \multicolumn{1}{c}{Tru} & \multicolumn{1}{c}{Rel} & \multicolumn{1}{c}{Out} & \multicolumn{1}{c}{Tho} & \multicolumn{1}{c}{Ner} & \multicolumn{1}{c}{Per} & \multicolumn{1}{c}{Hum}\\
Metric  & \multicolumn{16}{c}{$r$} \\ 
\Xhline{0.5\arrayrulewidth}
Majority & -0.041&-0.029&-0.104&-0.031&-0.122&-0.044&-0.065&-0.105&0.006&-0.077&-0.024&-0.085&-0.130&0.097&-0.127&-0.069 \\
SVM	& 0.063&0.086&-0.004&0.141&0.113&0.076&0.134&0.141&0.166&0.168&0.104&0.066&0.134&0.068&0.064&0.147\\
DF	& 0.240&0.273&0.017&0.139&0.112&0.173&0.118&0.217&0.148&0.143&0.019&0.093&0.041&0.136&0.168&0.259 \\
EF-LSTM			& 0.200&0.302&0.031&0.079&0.170&0.244&0.265&0.240&0.142&0.062&0.083&0.152&0.260&0.105&0.217&0.227\\
EF-SLSTM & 0.221&0.327&0.042&0.151&0.177&0.239&0.268&0.248&0.204&0.069&0.092&0.215&0.252&0.159&0.218&0.196 \\
EF-BLSTM & 0.162&0.289&-0.034&0.135&0.191&0.279&0.274&0.231&0.184&0.154&0.093&0.147&0.245&0.166&0.243&0.272\\
EF-SBLSTM  & 0.174&0.310&0.021&0.088&0.170&0.224&0.261&0.241&0.155&0.163&0.097&0.120&0.215&0.121&0.216&0.171\\
MV-LSTM	& 0.358&0.416&0.131&0.146&0.280&0.347&0.323&0.326&0.295&0.237&0.119&0.238&0.284&0.258&0.239&0.317 \\
BC-LSTM		& 0.359&0.425&0.081&0.234&0.358&0.417&0.450&0.361&0.293&0.109&0.075&0.078&0.363&0.184&0.344&0.319\\
TFN &  0.089&0.201&0.030&0.020&0.124&0.204&0.171&0.223&-0.051&-0.064&0.114&0.060&0.048&-0.002&0.106&0.213 \\ 
MARN & 0.340&0.410&0.166&0.235&0.340&0.374&0.406&0.378&0.282&0.147&0.215&0.204&0.348&0.235&0.303&0.287 \\
MFN  & {0.395}&{0.428}&{0.193}&{0.313}&{0.367}&{0.431}&{0.452}&{0.395}&{0.333}&{0.296}&{0.255}&{0.259}&{0.381}&{0.318}&{0.377}&{0.386}\\ 
\Xhline{0.5\arrayrulewidth} 
{\ours }  & \textbf{0.431}&\textbf{0.450}&\textbf{0.197}&\textbf{0.411}&\textbf{0.380}&\textbf{0.448}&\textbf{0.467}&\textbf{0.452}&\textbf{0.368}&0.212&\textbf{0.309}&\textbf{0.333}&\textbf{0.404}&\textbf{0.333}&0.334&\textbf{0.408} \\
%\Xhline{0.5\arrayrulewidth}
$\Delta_{SOTA}$	& \textcolor{gg}{$\uparrow $ {0.036}} & \textcolor{gg}{$\uparrow $ {0.022}} & \textcolor{gg}{$\uparrow $ {0.004}} & \textcolor{gg}{$\uparrow $ {0.097}} & \textcolor{gg}{$\uparrow $ {0.013}} & \textcolor{gg}{$\uparrow $ {0.017}} & \textcolor{gg}{$\uparrow $ {0.015}} & \textcolor{gg}{$\uparrow $ {0.057}} & \textcolor{gg}{$\uparrow $ {0.035}} & -- & \textcolor{gg}{$\uparrow $ {0.054}} & \textcolor{gg}{$\uparrow $ {0.074}} & \textcolor{gg}{$\uparrow $ {0.023}} & \textcolor{gg}{$\uparrow $ {0.015}} & -- & \textcolor{gg}{$\uparrow $ {0.022}} \\
\Xhline{3\arrayrulewidth}
\end{tabular}
\label{pom}
\end{table*}
\newcolumntype{K}[1]{>{\centering\arraybackslash}p{#1}}
%\definecolor{gg}{RGB}{45,190,45}
\definecolor{gg}{RGB}{0,0,0}

\begin{table*}[!htbp]
\fontsize{7}{10}\selectfont
%\centering
\caption{Sentiment prediction results on CMU-MOSI, ICT-MMMO, YouTube and MOUD. The best results are highlighted in bold and $\Delta_{SOTA}$ shows the change in performance over previous state of the art (SOTA). Improvements are highlighted in green. \ours \ achieves state-of-the-art or competitive performance on all datasets and metrics.}
\setlength\tabcolsep{0.7pt}
\begin{tabular}{l : *{5}{K{1.07cm}} : *{2}{K{1.07cm}} : *{2}{K{1.07cm}} : *{2}{K{1.07cm}}}
\Xhline{3\arrayrulewidth}
Dataset       & \multicolumn{5}{c:}{\textbf{CMU-MOSI}} & \multicolumn{2}{c:}{\textbf{ICT-MMMO}} & \multicolumn{2}{c}{\textbf{YouTube}} & \multicolumn{2}{c}{\textbf{MOUD}}  \\
Task			& \multicolumn{5}{c:}{Sentiment} & \multicolumn{2}{c:}{Sentiment} & \multicolumn{2}{c}{Sentiment} & \multicolumn{2}{c}{Sentiment} \\
Metric   & Acc\_$7$   & Acc\_$2$ & F1   & MAE & $r$ & Acc\_$2$ & F1 & Acc\_$3$ & F1 & Acc\_$2$ & F1 \\ 
\Xhline{0.5\arrayrulewidth}
Majority   & 17.5 & 50.2 &  50.1 &1.864 &  0.057  & 40.0 & 22.9 & 42.4 & 25.2 & 60.4 & 45.5\\
RF         & 21.3& 56.4 &  56.3 & - &  - &   70.0 & 69.8 & 33.3 & 32.3 & 64.2 & 63.3\\
SVM-MD     & 26.5 & 71.6 & 72.3 & 1.100  & 0.559 & 68.8 & 68.7 & 42.4 & 37.9 & 59.4 & 45.5\\
THMM	   & 17.8 & 53.8 & 53.0  & - & & 50.7	& 45.4	&  42.4 & 27.9 & 61.3 & 57.0\\
SAL-CNN    & -  & 73.0 &  - &   -      &  -	&  -  & - & - & - & - & - \\ 
C-MKL      &  30.2	 & 72.3 &   72.0   &  - & -    & - & - & - & - & - & - \\
EF-HCRF	   &  24.6	& 65.3 & 65.4  & - & - 	 & 50.0 & 50.3 & 44.1 & 43.8 & 54.7 & 54.7 \\
EF-LDHCRF  &  24.6	& 64.0 & 64.0 &  - & -	  & 73.8 & 73.1 & 45.8 & 45.0 & 52.8 & 49.3 \\
MV-HCRF	   &  22.6	& 44.8 & 27.7 & - & - 	 & 36.3 & 19.3 & 27.1 & 19.7 & 60.4 & 45.5 \\
MV-LDHCRF  &  24.6	& 64.0 & 64.0  & - & -	 & 68.8 & 67.1 & 44.1 & 44.0 & 53.8 & 46.9 \\
CMV-HCRF   &  22.3	& 44.8 & 27.7 & - & -	  & 36.3 & 19.3 & 30.5 & 14.3 & 60.4 & 45.5 \\
CMV-LDHCRF &  24.6	& 63.6 & 63.6 & - & - 	 & 51.3 & 51.4 & 42.4 & 42.0 & 53.8 & 47.8 \\
EF-HSSHCRF &  24.6  & 63.3 & 63.4 & - & -	  & 50.0 & 51.3 & 37.3 & 35.6 & 52.8 & 49.3 \\
MV-HSSHCRF &  24.6  & 65.6 & 65.7 &- & - 	& 62.5 & 63.1 & 44.1 & 44.0 & 47.2 & 46.4 \\
DF         &  26.8 & 72.3 & 72.1 & 1.143 & 0.518 & 65.0 & 58.7 & 45.8 & 32.0 & 67.0 & 67.1 \\
EF-LSTM    &  32.4 & 74.3 & 74.3 & 1.023 & 0.622 & 66.3 & 65.0 & 44.1 & 43.6 & 67.0 & 64.3 \\
EF-SLSTM   & 29.3 & 72.7 & 72.8 & 1.081 & 0.600 & 72.5 & 70.9 & 40.7 & 41.2 & 56.6 & 51.4 \\
EF-BLSTM   & 28.9 & 72.0 & 72.0 & 1.080 & 0.577	 & 63.8 & 49.6 & 42.4 & 38.1 & 58.5 & 58.9 \\
EF-SBLSTM  & 26.8 & 73.3 & 73.2 & 1.037 & 0.619 & 62.5 & 49.0 & 37.3 & 33.2 & 63.2 & 63.3 \\
MV-LSTM	   & 33.2 & 73.9 & 74.0 & 1.019 & 0.601 & 72.5 & 72.3 & 45.8 & 43.3 & 57.6 & 48.2 \\
BC-LSTM    & 28.7 & 73.9 & 73.9 & 1.079 & 0.581 & 70.0 & 70.1 & 45.0 & 45.1 & 72.6 & 72.9 \\ 
TFN        & 28.7 & 74.6 & 74.5 & 1.040 & 0.587 & 72.5 & 72.6 & 45.0 & 41.0 & 63.2 & 61.7 \\ 
MARN & 34.7 & 77.1  & 77.0 & 0.968  & 0.625  & 71.3 & 70.2 & 48.3 & 44.9 & 81.1 & 81.2 \\ 
MFN & 34.1 & 77.4 & 77.3 & {0.965} & {0.632} & 73.8 & 73.1 & 51.7 & 51.6 & 81.1 & 80.4 \\
\Xhline{0.5\arrayrulewidth}
{\ours} & \textcolor{gg}{\textbf{36.2}}   & \textbf{78.1} & \textbf{78.1} & \textbf{0.951}	& \textcolor{gg}{\textbf{0.662}} & \textcolor{gg}{\textbf{81.3}} & \textcolor{gg}{\textbf{79.2}} & \textcolor{gg}{\textbf{53.3}} & \textcolor{gg}{\textbf{52.4}} & \textbf{82.1} & \textbf{81.7} \\
$\Delta_{SOTA}$	& \textcolor{gg}{$\uparrow $ {1.5}} & \textcolor{gg}{$\uparrow $ {0.7}} & \textcolor{gg}{$\uparrow $ {0.8}} & \textcolor{gg}{$\downarrow $ {0.014}} & \textcolor{gg}{$\uparrow $ {0.030}} & \textcolor{gg}{$\uparrow $ {7.5}} & \textcolor{gg}{$\uparrow $ {6.1}} & \textcolor{gg}{$\uparrow $ {1.6}} & \textcolor{gg}{$\uparrow $ {0.8}} & \textcolor{gg}{$\uparrow $ {1.0}} & \textcolor{gg}{$\uparrow $ {0.5}}  \\ 
\Xhline{3\arrayrulewidth}
\end{tabular}
\label{mosi}
\end{table*}
\newcolumntype{K}[1]{>{\centering\arraybackslash}p{#1}}
%\definecolor{gg}{RGB}{45,190,45}
\definecolor{gg}{RGB}{0,0,0}

\begin{table}[tb]
\fontsize{7}{10}\selectfont
%\centering
\caption{Emotion recognition results on IEMOCAP test set. The best results are highlighted in bold and $\Delta_{SOTA}$ shows the change in performance over previous SOTA. Improvements are highlighted in green. \ours \ achieves state-of-the-art or competitive performance on all datasets and metrics.}
\setlength\tabcolsep{0.71pt}
\vspace{2mm}
\begin{tabular}{l  *{12}{K{0.97cm}}}
\Xhline{3.0\arrayrulewidth}
Dataset & \multicolumn{12}{c}{\textbf{IEMOCAP Emotions}} \\
Task		& \multicolumn{2}{c}{Happy} & \multicolumn{2}{c}{Sad} & \multicolumn{2}{c}{Angry} & \multicolumn{2}{c}{Frustrated} & \multicolumn{2}{c}{Excited} & \multicolumn{2}{c}{Neutral} \\
Metric		& Acc\_$2$ & F1 & Acc\_$2$ & F1 & Acc\_$2$ & F1 & Acc\_$2$ & F1 & Acc\_$2$ & F1 & Acc\_$2$ & F1 \\
\Xhline{0.5\arrayrulewidth}
Majority		& 85.6 & 79.0 & 79.4 & 70.3 & 75.8 & 65.4 & 79.5 & 70.4 & 89.6 & 84.7 & 59.1 & 44.0 \\
SVM     		& 86.1 & 81.5 & 81.1 & 78.8 & 82.5 & 82.4 & 77.3 & 71.1 & 86.4 & 86.0 & 65.2 & 64.9 \\
RF     			& 85.5 & 80.7 & 80.1 & 76.5 & 81.9 & 82.0 & 78.6 & 75.3 & 88.9 & 85.1 & 63.2 & 57.3 \\
THMM			& 85.6 & 79.2 & 79.5 & 79.8 & 79.3 & 73.0 & 71.6 & 69.6 & 86.0 & 84.6 & 58.6 & 46.4 \\
EF-HCRF			& 85.7 & 79.2 & 79.4 & 70.3 & 75.8 & 65.4 & 79.5 & 70.4 & 89.6 & 84.7 & 59.1 & 44.0 \\
EF-LDHCRF		& 85.8 & 79.5 & 79.4 & 70.3 & 75.8 & 65.4 & 79.5 & 70.4 & 89.6 & 84.7 & 59.1 & 44.0 \\
MV-HCRF			& 15.0 & 4.9  & 79.4 & 70.3 & 24.2 & 9.4  & 79.5 & 70.4 & 89.6 & 84.7 & 59.1 & 44.0 \\
MV-LDHCRF		& 85.7 & 79.2 & 79.4 & 70.3 & 75.8 & 65.4 & 79.5 & 70.4 & 89.6 & 84.7 & 59.1 & 44.0 \\
CMV-HCRF		& 14.4 & 3.6  & 79.4 & 70.3 & 24.2 & 9.4  & 79.5 & 70.4 & 89.6 & 84.7 & 59.1 & 44.0 \\
CMV-LDHCRF		& 85.8 & 79.5 & 79.4 & 70.3 & 75.8 & 65.4 & 79.5 & 70.4 & 89.6 & 84.7 & 59.1 & 44.0 \\
EF-HSSHCRF		& 85.8 & 79.5 & 79.4 & 70.3 & 75.8 & 65.4 & 79.5 & 70.4 & 89.6 & 84.7 & 59.1 & 44.0 \\
MV-HSSHCRF		& 85.8 & 79.5 & 79.4 & 70.3 & 75.8 & 65.4 & 79.5 & 70.4 & 89.6 & 84.7 & 59.1 & 44.0 \\
DF   			& 86.0 & 81.0 & 81.8 & 81.2 & 75.8 & 65.4 & 78.4 & 76.8 & 89.6 & 84.7 & 59.1 & 44.0 \\
EF-LSTM   		& 85.2 & 83.3 & 82.1 & 81.1 & 84.5 & 84.3 & 79.5 & 70.4 & 89.6 & 84.7 & 68.2 & 67.1 \\
EF-SLSTM		& 85.6 & 79.0 & 80.7 & 80.2 & 82.8 & 82.2 & 77.5 & 69.7 & 89.3 & 86.2 & 68.8 & \textbf{68.5} \\
EF-BLSTM 		& 85.0 & 83.7 & 81.8 & 81.6 & 84.2 & 83.3 & 79.5 & 70.4 & 89.6 & 84.7 & 67.1 & 66.6 \\
EF-SBLSTM 		& 86.0 & 84.2 & 80.2 & 80.5 & {85.2} & {84.5} & 79.5 & 70.4 & 89.6 & 84.7 & 67.8 & 67.1 \\
MV-LSTM   		& 85.9 & 81.3 & 80.4 & 74.0 & 85.1 & 84.3 & 79.5 & 73.8 & 88.9 & 85.8 & 67.0 & 66.7 \\
BC-LSTM    		& 84.9 & 81.7 & 83.2 & 81.7 & 83.5 & 84.2 & 80.0 & 76.1 & 86.9 & 85.4 & 67.5 & 64.1 \\
TFN      		& 84.8 & 83.6 & 83.4 & {82.8} & 83.4 & 84.2 & 74.1 & 74.3 & 75.6 & 78.0 & 67.5 & 65.4 \\
MARN			& 86.7 & 83.6 & 82.0 & 81.2 & 84.6 & 84.2 & 79.5 & 76.6 & 89.6 & \textbf{87.1} & 66.8 & 65.9 \\
MFN				& 90.1 & 85.3 & 85.8 & 79.2 & 87.0 & 86.0 & 80.3 & \textbf{76.9} & 89.8 & 86.3 & 71.8 & 61.7 \\
\Xhline{0.5\arrayrulewidth}
{\ours}      & \textbf{90.2} & \textbf{85.8} & \textbf{88.4} & \textbf{86.1} & \textbf{87.5} & \textbf{86.7} & \textbf{80.4} & 74.5 & \textbf{90.0} & \textbf{87.1} & \textbf{72.1} & 68.1 \\
$\Delta_{SOTA}$	& \textcolor{gg}{$\uparrow $ {0.1}} & \textcolor{gg}{$\uparrow $ {0.5}} & \textcolor{gg}{$\uparrow $ {2.6}} & \textcolor{gg}{$\uparrow $ {3.3}} & \textcolor{gg}{$\uparrow $ {0.5}} & \textcolor{gg}{$\uparrow $ {0.7}} & \textcolor{gg}{$\uparrow $ {0.1}} & -- & \textcolor{gg}{$\uparrow $ {0.2}} & -- & \textcolor{gg}{$\uparrow $ {0.3}} & -- \\
\Xhline{3\arrayrulewidth}
\end{tabular}
\label{iemocap}
\end{table}

\iffalse
\section{Choice of Encoder $Q(\cdot, \cdot)$}

In our objective (i.e., Eq. \eqref{eq:approxmulti}), we have the freedom of choosing any form of the encoder $Q(\cdot, \cdot)$. In the paper, we choose a {\em Gaussian encoder} for $Q(\mathbf{Z|X})$ with reparametrization trick \citep{kingma2013auto}, which means we map a data $x \in \mathcal{X}$ to a distribution $Q(\mathbf{\mathbf{Z}|\mathbf{X}} = x)$:
\begin{equation}
\label{eq:Q}
Q(\mathbf{Z|X}=x) = \mu(x) + \epsilon \cdot \sigma(x)\,\,\,\mathrm{with}\,\,\,\epsilon \sim \mathcal{N}(0, \mathbf{I}). 
\end{equation} 
Note that the {\em Gaussian encoder} degenerates to the deterministic one when we force $\epsilon = 0$.

The advantage of choosing a stochastic encoder over a deterministic encoder for $Q(\mathbf{Z|X})$ is to mitigate the problems of the mismatch between the dimension $d_{\mathbf{Z}}$ of the latent variable and the intrinsic dimension $d_{I}$ of the dataset \citep{rubenstein2018latent}. For example, there would be many {\em holes} in the latent space if we use a deterministic encoder.
\fi

\section{Information and Gradient-Based Interpretation}

{\bf Information-Based Interpretation:} We choose the normalized Hilbert-Schmidt Independence Criterion~\citep{gretton2005measuring,wu2018dependency} as the approximation (see~\cite{sugiyama2012kernel,wu2018dependency}) of our MI measure:
\begin{equation}
\small
\label{eq:HSIC}
\textnormal{MI}(\mathbf{F_\cdot}, \mathbf{\hat{X}}_i) = \mathrm{HSIC}_{norm}(\mathbf{F_\cdot}, \mathbf{\hat{X}}_i) = \frac{\mathrm{tr}(\mathbf{K_{F_\cdot}}\mathbf{H}\mathbf{K}_{\mathbf{\hat{X}}_i} \mathbf{H})}{\|\mathbf{H}\mathbf{K_{F_\cdot}}\mathbf{H}\|_F\|\mathbf{H}\mathbf{K}_{\mathbf{\hat{X}}_i}\mathbf{H}\|_F},
\end{equation}
where $\cdot$ represents $y$ or $a_{i}$, $n$ is the number of \{$\mathbf{F_\cdot}, \mathbf{\hat{X}}_i$\} pairs, $\mathbf{H} = \mathbf{I} - \frac{1}{n}\mathbf{1}\mathbf{1}^\top$, $\mathbf{K_{F_\cdot}} \in \mathbb{R}^{n\times n}$ is the Gram matrix of $\mathbf{F_\cdot}$ with $\mathbf{K_{F_\cdot}}_{ij} = k_1(\mathbf{F_\cdot}_{i}, \mathbf{F_\cdot}_{j})$, $\mathbf{K}_{\mathbf{\hat{X}}_i} \in \mathbb{R}^{n\times n}$ is the Gram matrix of $\mathbf{\hat{X}}_i$ with ${{{\mathbf{K}}_{\mathbf{\hat{X}}}}_i}_{jk} = k_2({\mathbf{\hat{X}}}_{ij}, {\mathbf{\hat{X}}}_{ik})$. $k_1(\cdot, \cdot)$ and $k_2(\cdot, \cdot)$ are predefined kernel functions.

The most common choice for the kernel is the RBF kernel. However, if we consider time series data with various time steps, we need to either perform data augmentation or choose another kernel choice. For example, we can adopt the Global Alignment Kernel~\citep{cuturi2007kernel} which considers the alignment between two varying-length time series when computing the kernel score between them. To simplify our analysis, we choose to augment data before we calculate the kernel score with the RBF kernel. More specifically, we perform averaging over time series data:
\begin{equation}
\mathbf{X}_{aug} = \frac{1}{n} \sum_{t=1}^T X^t\,\,\mathrm{with}\,\, \mathbf{X} = [X^1, X^2, \cdots , X^T].
\end{equation}
The bandwidth of the RBF kernel is set as 1.0 throughout the experiments.

\begin{center}
\begin{table}[h]
\caption{Information-Based interpretation results showing ratios $r_i = \frac{\textrm{MI}(\mathbf{F_y},\hat{\mathbf{X}}_{i})}{\textrm{MI}(\mathbf{F_a}_{i},\hat{\mathbf{X}}_{i})}$, $i \in \{ (\ell)anguage, (v)isual, (a)coustic \} $ for the POM dataset for personality traits prediction.}
\fontsize{7}{10}\selectfont
\centering
\setlength\tabcolsep{5pt}
\vspace{-2mm}
\begin{tabular}{| c || *{3}{K{1.5cm}}|}
\hline
Ratio		& $r_\ell$ (language) & $r_v$ (visual) & $r_a$ (acoustic) \\ 
\hline \hline
POM		& 1.090 & 0.996 & 0.898 \\
\hline
\end{tabular}
\label{tbl:HSIC}
\vspace{-10mm}
\end{table}
\end{center}

Here, we provide an additional interpretation result for the POM dataset in Table~\ref{tbl:HSIC}. We observe that the language modality is also the most informative while the visual and acoustic modalities are almost equally informative. This result is in agreement with behavioral studies which have observed that non-verbal behaviors are particularly informative of personality traits~\citep{10.1371/journal.pone.0037450,Levine:2009:RPS:1618452.1618518,personality}. For example, the same sentence ``this movie was great'' can convey significantly different messages on speaker confidence depending on whether it was said in a loud and exciting voice, with eye contact, or powerful gesticulation.

{\bf Gradient-Based Interpretation:} \ours \ reconstructs $x_i$ as follows:
\begin{equation}
\small
\label{eq:grad_time_1}
\begin{split}
\hat{x}_i = F_i (f_{ai},f_y), f_{ai} = G_{ai}(z_{ai}), f_y = G_y(z_y) ,  z_{ai} \sim Q(\mathbf{Z}_{\mathbf{a}i}|\mathbf{X}_{i} = x_{i}), z_y \sim Q(\mathbf{Z_y}|\mathbf{X}_{1:M} = x_{1:M}).
\end{split}
\end{equation}

Equation~\eqref{eq:grad_time_1} also explains how we obtain $f_y \sim P(\mathbf{F_y}|\mathbf{X}_{1:M} = x_{1:M})$. The gradient flow through time is defined as:
\begin{equation}
\label{eq:grad_time_2}
\begin{split}
%\nabla_{f_{ai}} (\hat{x}_i) := & [\|\nabla_{f_{ai}} \hat{x}_i^1\|_F^2, \|\nabla_{f_{ai}} \hat{x}_i^2\|_F^2, \cdots, \|\nabla_{f_{ai}} \hat{x}_i^T\|_F^2] \\ 
\nabla_{f_{y}} (\hat{x}_i) := & [\|\nabla_{f_{y}} \hat{x}_i^1\|_F^2, \|\nabla_{f_{y}} \hat{x}_i^2\|_F^2, \cdots, \|\nabla_{f_{y}} \hat{x}_i^T\|_F^2].
\end{split}
\end{equation}

\section{Encoder and Decoder Design for Multimodal Synthetic Image Dataset}

For experiments on the multimodal synthetic image dataset, we use convolutional+fully-connected layers for the encoder and deconvolutional+fully-connected layers for the decoder~\citep{Zeiler2010DeconvolutionalN}. Different convolutional layers are each applied on the input SVHN and MNIST images to learn modality-specific generative factors. Next, we concatenate the features from two more convolutional layers on SVHN and MNIST to learn the multimodal-discriminative factor. The multimodal discriminative factor is passed through fully-connected layers to predict the label. For generation, we concatenate the multimodal discriminative factors and the modality-specific generative factor together and use a deconvolutional layer to generate digits.

\section{Encoder and Decoder Design for Multimodal Time Series Datasets}

\begin{figure*}[h]
	\centering
	\vspace{-5mm}
	\includegraphics[width=0.9\textwidth]{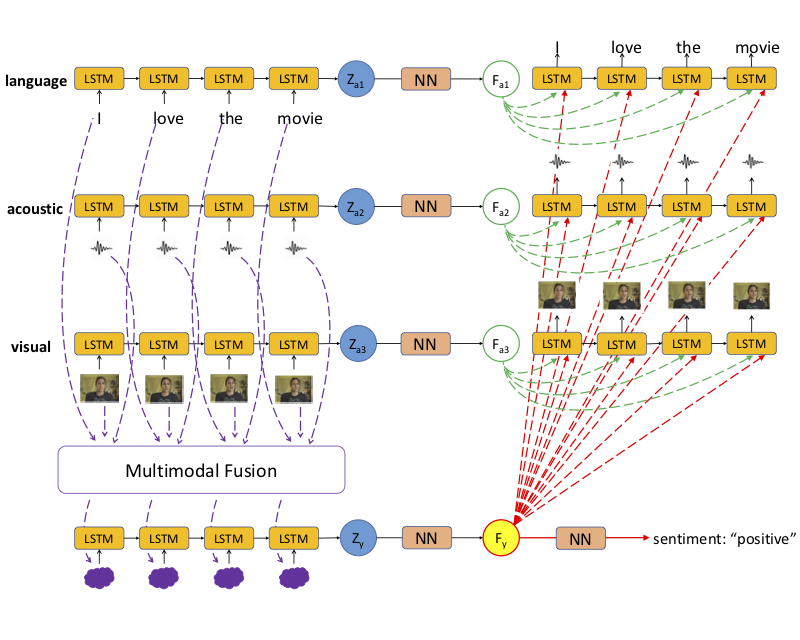}
	\vspace{-8mm}
	\caption{Recurrent neural architecture for \ours. The encoder $Q(\mathbf{Z_y}|\mathbf{X}_{1:M})$ can be parametrized by any model that performs multimodal fusion~\citep{Nojavanasghari:2016:DMF:2993148.2993176,zadeh2018memory}. We use encoder LSTM networks and decoder LSTM networks~\citep{cho-al-emnlp14} to parametrize functions $Q(\mathbf{Z_a}_{1:M}|\mathbf{X}_{1:M})$ and $F_{1:M}$ respectively, and FCNNs to parametrize functions $G_y$, ${G_a}_{\{1:M\}}$ and $D$.}
	\label{fig:illus_LSTM}
\end{figure*}

Figure~\ref{fig:illus_LSTM} illustrates how \ours \ operates on multimodal time series data. The encoder $Q(\mathbf{Z_y}|\mathbf{X}_{1:M})$ can be parametrized by any model that performs multimodal fusion~\citep{Nojavanasghari:2016:DMF:2993148.2993176,zadeh2018memory}. We choose the Memory Fusion Network (MFN)~\citep{zadeh2018memory} as our encoder $Q(\mathbf{Z_y}|\mathbf{X}_{1:M})$. We use encoder LSTM networks and decoder LSTM networks~\citep{cho-al-emnlp14} to parametrize functions $Q(\mathbf{Z_a}_{1:M}|\mathbf{X}_{1:M})$ and $F_{1:M}$ respectively, and FCNNs to parametrize functions $G_y$, ${G_a}_{\{1:M\}}$ and $D$.

\section{Surrogate Inference Graphical Model}
\begin{figure*}[h]
	\centering
	\includegraphics[width=0.2\textwidth]{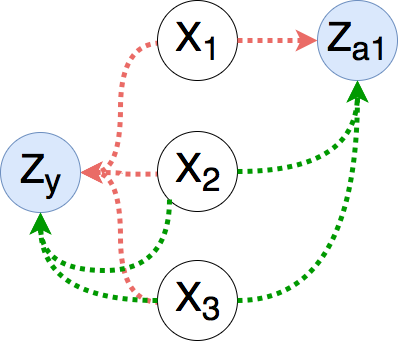}
	%\vspace{-5mm}
	\caption{The surrogate inference graphical model to deal with missing modalities in \ours. Red lines denote original inference in \ours \ and green lines denote surrogate inference to infer latent codes given present modalities.}
	\label{fig:missing}
\end{figure*}

We illustrate the surrogate inference for addressing the missing modalities issue in Figure~\ref{fig:missing}. The surrogate inference model infers the latent codes given the present modalities. These inferred latent codes can then be used for reconstructing the missing modalities or label prediction in the presence of missing modalities.

\section{Comparison with~\cite{hsu2018disentangling}}

A similar approach for factorizing the latent factors was recently proposed by~\cite{hsu2018disentangling} in work that was performed independently and in parallel. In comparison with \ours, there are several major differences that can be categorized into the {(1)} prior matching discrepancy, {(2)} inference network, {(3)} discriminative objective, {(4)} multimodal fusion, {(5)} scale of experiments.

\begin{enumerate}
\item \ours \ uses $\mathcal{MMD}(Q_{\mathbf{Z}}, P_{\mathbf{Z}})$ (see Equation~\ref{eq:approxmulti}) as the prior matching discrepancy while~\cite{hsu2018disentangling} use $\mathcal{KL}(Q_{\mathbf{Z|X}}, P_{\mathbf{Z}})$) (see Section 2.2.1 in~\cite{hsu2018disentangling}).
\item \ours \ considers multimodal and unimodal inference in a single network (see Figure~\ref{fig:illus}(b)), while~\cite{hsu2018disentangling} considers separate networks (see Figure 1 in \cite{hsu2018disentangling}). They further propose to match the coherence between these two networks using an additional loss term (see Equation 7 in~\cite{hsu2018disentangling}).
\item \ours \ learns to predict the labels using a generative framework (see Figure~\ref{fig:illus}(a)), while~\cite{hsu2018disentangling} use an additional hinge loss to separate the latent factors from different labels (see Equation 9 in~\cite{hsu2018disentangling}).
\item \ours \ is a flexible framework that can be combined with any multimodal fusion encoder (see Section~\ref{enc_dec}), while~\cite{hsu2018disentangling} considers a fixed multimodal encoder (similar to early fusion) (see Section 4.1 in~\cite{hsu2018disentangling}).
\item We evaluate the performance of \ours \ over a much larger scale of datasets. We perform experiments on six multimodal time-series datasets that take on the form of videos with the language, visual, and acoustic modalities. These datasets span three core research areas of multimodal personality traits recognition, multimodal sentiment analysis, and multimodal emotion recognition. On the other hand,~\cite{hsu2018disentangling} evaluates their model on a spoken digit dataset which randomly combines a digit image with a spoken digit (see Section 4 in~\cite{hsu2018disentangling}). \ours \ further considers experiments to evaluate reconstruction and prediction in the presence of missing modalities (see Section~\ref{disc}) which~\cite{hsu2018disentangling} do not. Lastly, we compares to over 20 baseline models in our experiments (see Section~\ref{disc}) and explore the choice of various multimodal encoders in \ours. \cite{hsu2018disentangling} only compares to the JMVAE baseline model~\citep{Suzuki2016JointML} which resembles the $\mathbf{M_D}$ model in our ablation study (see Section 4.4 in~\cite{hsu2018disentangling}).
\end{enumerate}

In Table~\ref{compare_hsu}, we provide a comparison on the CMU-MOSI, ICT-MMMO, YouTube and MOUD datasets to test the disentanglement and prediction performance for the model described in~\cite{hsu2018disentangling}. These experimental results show that across these datasets and metrics, \ours \ performs better than the model proposed in~\cite{hsu2018disentangling}.  We would like to highlight that at the time of submission, the code for~\citep{hsu2018disentangling} had not been made public and we reimplemented their model to experiment on our datasets.

\newcolumntype{K}[1]{>{\centering\arraybackslash}p{#1}}
\definecolor{gg}{RGB}{0,0,0}
\begin{table*}[t!] %[!htbp]
\vspace{-5mm}
\caption{Comparison with \cite{hsu2018disentangling} for sentiment analysis on CMU-MOSI, ICT-MMMO, YouTube, and MOUD. \ours \ outperforms the baselines across these datasets and metrics.}

\fontsize{8}{12}\selectfont
\centering
\setlength\tabcolsep{0.7pt}
\begin{tabular}{|c || *{5}{K{0.88cm}} || *{2}{K{0.88cm}} || *{2}{K{0.88cm}} || *{2}{K{0.88cm}}|}
\hline
Dataset & \multicolumn{5}{c||}{{CMU-MOSI}} & \multicolumn{2}{c||}{{ICT-MMMO}} & \multicolumn{2}{c||}{{YouTube}} & \multicolumn{2}{c|}{{MOUD}}  \\
Task			& \multicolumn{5}{c||}{Sentiment} & \multicolumn{2}{c||}{Sentiment} & \multicolumn{2}{c||}{Sentiment} & \multicolumn{2}{c|}{Sentiment} \\
Metric  & Acc\_$7$ & Acc\_$2$ & F1 & MAE & $r$ & Acc\_$2$ & F1 & Acc\_$3$ & F1 & Acc\_$2$ & F1 \\ 
\hline \hline
without factorization (EF)    & 32.4  &  74.3  &  74.3 &  1.023   &  0.622 & 72.5 & 70.9 & 44.1 & 43.6 & 67.0 & 64.3 \\ 
without factorization (MFN)   & 34.1 &  77.4  &  77.3 &  0.965 & 0.632 & 73.8 & 73.1 & 51.7 & 51.6 & 81.1 & 80.4 \\ \hline \hline
\cite{hsu2018disentangling}  & 33.8 & 75.2 & 75.2 & 1.049 & 0.584 & 77.5 & 75.0 & 51.7 & 48.6 & 66.0 & 62.9 \\ \hline \hline
\ours \ & \textcolor{gg}{\textbf{36.2}}   & \textbf{78.1} & \textbf{78.1} & \textbf{0.951}	& \textcolor{gg}{\textbf{0.662}} & \textcolor{gg}{\textbf{81.3}} & \textcolor{gg}{\textbf{79.2}} & \textcolor{gg}{\textbf{53.3}} & \textcolor{gg}{\textbf{52.4}} & \textbf{82.1} & \textbf{81.7} \\
\hline
\end{tabular}
\label{compare_hsu}
\end{table*}

\section{Comparison with $\beta$-VAE}

Although $\beta$-VAE~\citep{beta_vae} was designed to handle unimodal data, we provide an extension to multimodal data. To achieve this, we set the choice of prior matching discrepancy as the KL-divergence $\mathcal{KL}(Q_{\mathbf{Z|X}}, P_{\mathbf{Z}})$ and set $\beta$ large (i.e. $\beta \in \{10,50,100,200\}$) to encourage disentanglement of latent variables. We train a $\beta$-VAE to model multimodal data using the same factorization as proposed in our model (i.e. modality-specific generative factors $\mathbf{Z_a}_{\{1:M\}}$ and a multimodal discriminative factor $\mathbf{Z_y}$). To provide a fair comparison to our discriminative model, we fine tune by training a classifier on top of the multimodal discriminative factor $\mathbf{Z_y}$ to the label $\mathbf{Y}$. We provide experimental results in Table~\ref{compare_vae} on the CMU-MOSI, ICT-MMMO, YouTube and MOUD datasets. \ours \ outperforms $\beta$-VAE across these datasets and metrics.

\newcolumntype{K}[1]{>{\centering\arraybackslash}p{#1}}
\definecolor{gg}{RGB}{0,0,0}
\begin{table*}[t!] %[!htbp]
\caption{Comparison with $\beta$-VAE for multimodal sentiment analysis on CMU-MOSI, ICT-MMMO, YouTube, and MOUD. \ours \ outperforms $\beta$-VAE across these datasets and metrics.}

\fontsize{8}{12}\selectfont
%\centering
\setlength\tabcolsep{0.7pt}
\begin{tabular}{|c || *{5}{K{1.09cm}} || *{2}{K{1.09cm}} || *{2}{K{1.09cm}} || *{2}{K{1.09cm}}|}
\hline
Dataset       & \multicolumn{5}{c||}{{CMU-MOSI}} & \multicolumn{2}{c||}{{ICT-MMMO}} & \multicolumn{2}{c||}{{YouTube}} & \multicolumn{2}{c|}{{MOUD}}  \\
Task			& \multicolumn{5}{c||}{Sentiment} & \multicolumn{2}{c||}{Sentiment} & \multicolumn{2}{c||}{Sentiment} & \multicolumn{2}{c|}{Sentiment} \\
Metric  & Acc\_$7$ & Acc\_$2$ & F1 & MAE & $r$ & Acc\_$2$ & F1 & Acc\_$3$ & F1 & Acc\_$2$ & F1 \\ 
\hline \hline
$\beta$-VAE    & 29.7 & 71.3 & 71.3 & 1.094 & 0.552 & 65.0 & 59.8 & 46.7 & 31.3 & 60.4 & 54.2 \\
{\ours} & \textcolor{gg}{\textbf{36.2}}   & \textbf{78.1} & \textbf{78.1} & \textbf{0.951}	& \textcolor{gg}{\textbf{0.662}} & \textcolor{gg}{\textbf{81.3}} & \textcolor{gg}{\textbf{79.2}} & \textcolor{gg}{\textbf{53.3}} & \textcolor{gg}{\textbf{52.4}} & \textbf{82.1} & \textbf{81.7} \\
\hline
\end{tabular}

\label{compare_vae}
\vspace{-4mm}
\end{table*}

\end{document}